\documentclass[10pt,journal,twocolumn,twoside,compsoc]{IEEEtran}

\usepackage{url}
\usepackage[american]{babel}

\usepackage[caption=false]{subfig}
\usepackage{graphicx}
\usepackage{amsmath,amssymb} 
\usepackage{algorithm,algorithmic}
\usepackage{changepage}
\usepackage{multirow}
\usepackage[compress]{cite}

\newcommand{\squishlist}{
 \begin{list}{$\bullet$}
  { \setlength{\itemsep}{0pt}
     \setlength{\parsep}{1pt}
     \setlength{\topsep}{1pt}
     \setlength{\partopsep}{0pt}
     \setlength{\leftmargin}{1.5em}
     \setlength{\labelwidth}{1em}
     \setlength{\labelsep}{0.5em} } }

\newcommand{\squishend}{
  \end{list}  }

\usepackage[pagebackref=false,breaklinks=true,colorlinks,bookmarks=false]{hyperref}

\begin{document}
\title{A Tube-and-Droplet-based Approach for Representing and Analyzing Motion Trajectories}

\author{Weiyao Lin,~Yang Zhou,~Hongteng Xu,~Junchi Yan,~Mingliang Xu,~Jianxin Wu,~and Zicheng Liu%
\IEEEcompsocitemizethanks{
\IEEEcompsocthanksitem W. Lin and Y. Zhou are with the Dept. of Electronic Engineering, Shanghai Jiao Tong University, Shanghai, China. E-mail: \{wylin, zhz128ly\}@sjtu.edu.cn.
\IEEEcompsocthanksitem H. Xu is with the Dept. of Electrical and Computer Engineering, Georgia Institute of Technology, Atlanta, USA. E-mail: hxu42@gatech.edu.
\IEEEcompsocthanksitem J. Yan is with IBM Research, Shanghai, China. E-mail: yanjc@cn.ibm.com.
\IEEEcompsocthanksitem M. Xu is with Zhengzhou University, Zhengzhou, China. E-mail: iexumingliang@zzu.edu.cn.
\IEEEcompsocthanksitem J. Wu is with the National Key Laboratory for Novel Software Technology, Nanjing University, Nanjing, China. E-mail: wujx2001@nju.edu.cn.
\IEEEcompsocthanksitem Z. Liu is with Microsoft Research, Redmond, USA. E-mail: zliu@microsoft.com.
}
}

\maketitle

\begin{abstract}
Trajectory analysis is essential in many applications. In this paper, we address the problem of representing motion trajectories in a highly informative way, and consequently utilize it for analyzing trajectories. Our approach first leverages the complete information from given trajectories to construct a thermal transfer field which provides a context-rich way to describe the global motion pattern in a scene. Then, a 3D tube is derived which depicts an input trajectory by integrating its surrounding motion patterns contained in the thermal transfer field. The 3D tube effectively: 1) maintains the movement information of a trajectory, 2) embeds the complete contextual motion pattern around a trajectory, 3) visualizes information about a trajectory in a clear and unified way. We further introduce a droplet-based process. It derives a droplet vector from a 3D tube, so as to characterize the high-dimensional 3D tube information in a simple but effective way. Finally, we apply our tube-and-droplet representation to trajectory analysis applications including trajectory clustering, trajectory classification \& abnormality detection, and 3D action recognition. Experimental comparisons with state-of-the-art algorithms demonstrate the effectiveness of our approach. Code for our work is available at \url{http://min.sjtu.edu.cn/lwydemo/Trajectory analysis.htm}
\end{abstract}


\section{Introduction}

\IEEEPARstart{M}{otion} information, which reflects the temporal variation of visual contents, is essential in depicting the semantic contents in videos. As the motion information of many semantic contents is described by motion trajectories, trajectory analysis is of considerable importance to many applications including video surveillance, object behavior analysis, and video retrieval \cite{2,3,4,10,58}. Formally, trajectory analysis can be defined as the problem of deciding the class of one or more input trajectories according to their shapes and motion routes \cite{10,11,17,32}.

A motion trajectory is in general obtained by tracking an object over frames and linking object positions into a position sequence \cite{8,2,3}. Although trajectories contain detailed information of object movements, reliable trajectory analysis remains challenging due to the uncertain nature of object motion and the ambiguity from similar motion patterns.

\begin{figure}
  \centering
  \subfloat[]{\includegraphics[height = 2.23 cm, width =3.2 cm]{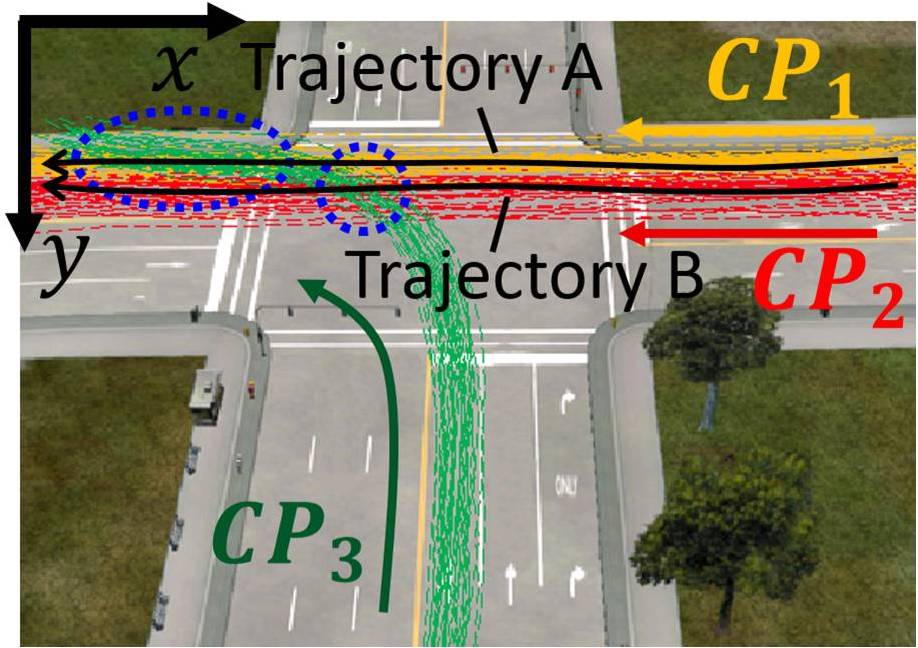}\label{fig:similar_example_a}}
   \hspace{0mm}
  \subfloat[]{\includegraphics[height = 1.88 cm, width = 2.5 cm]{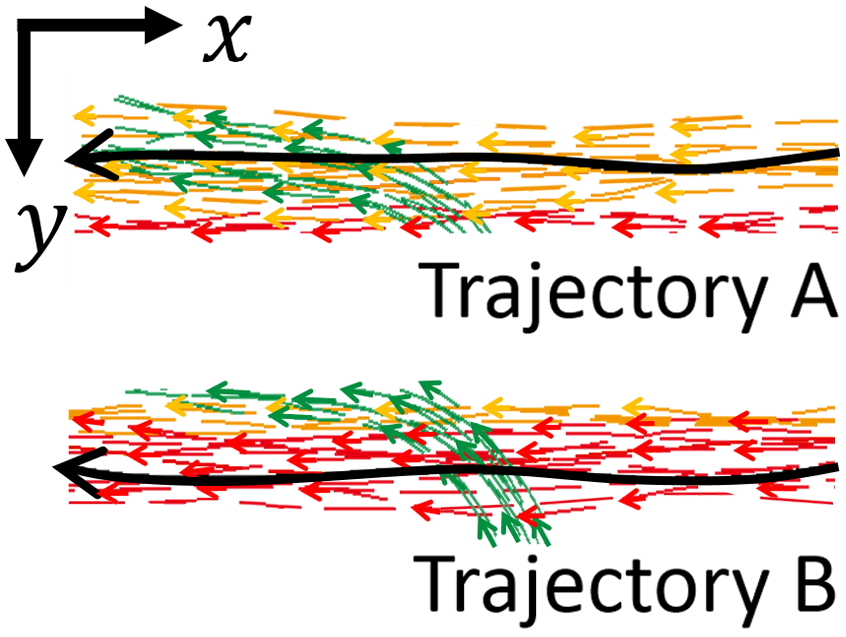}\label{fig:similar_example_b}}
  \hspace{0mm}
  \subfloat[]{\includegraphics[height = 1.88 cm, width =2.9 cm]{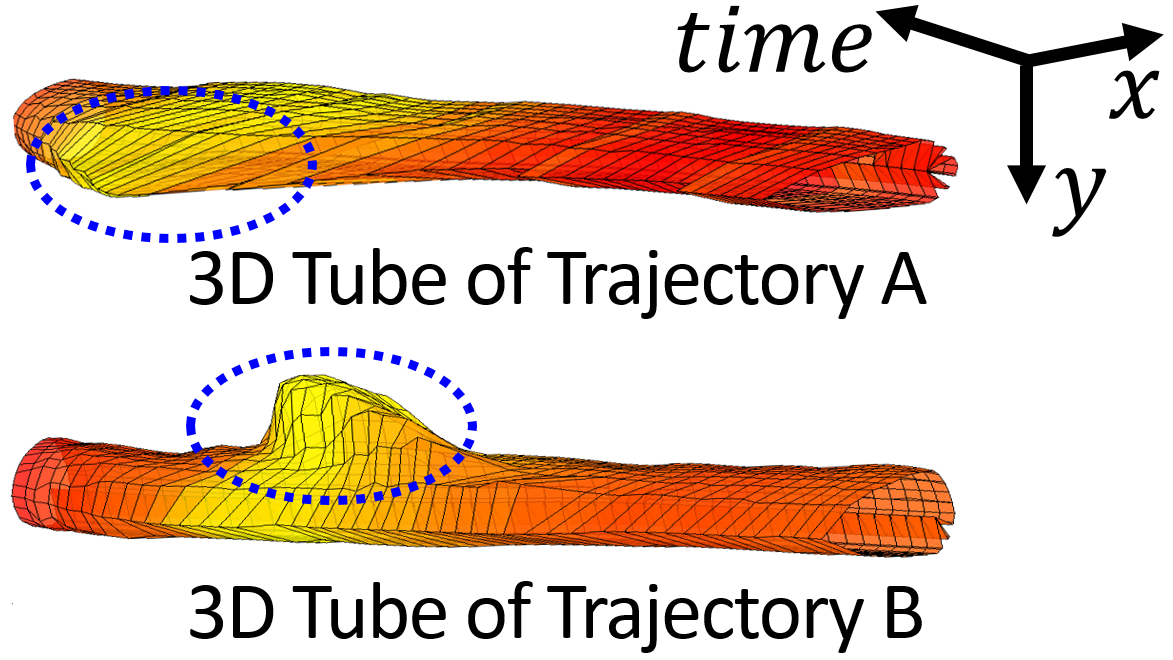}\label{fig:similar_example_c}}
  \vspace{-1.5mm}
  \caption{(a) An example of ambiguous trajectories: The orange, red, and green curves labeled by $CP_1$-$CP_3$ indicate three trajectory classes. The black curves labeled by \emph{Trajectory} $A$ and $B$ are two trajectories from $CP_1$ and $CP_2$, respectively. (b) Complete contextual motion patterns of trajectories $A$ and $B$ in (a): Contextual motion patterns are described by the motion flows of all trajectory points in the neighborhood of $A$ and $B$ (Note: We use color to differentiate motion flows from different trajectory classes only for a clearer illustration. In our approach, we do not differentiate motion flows' classes and directly use \emph{all} motion flows in the neighborhood of an input trajectory to model its contextual motion pattern).
  (c) Use our 3D tube to model $A$ and $B$'s complete contextual motion patterns. (Best viewed in color)}\label{fig:similar_example}
\end{figure}

One major challenge for trajectory analysis is to differentiate trajectory classes with only subtle differences. For example, Fig.~\ref{fig:similar_example_a} shows three trajectory classes $CP_1$, $CP_2$, and $CP_3$, where $CP_1$ and $CP_2$ include vehicle trajectories following two adjacent leftward street lanes and $CP_3$ includes vehicle trajectories following a left-turn street lane.
Since trajectories in $CP_1$ and $CP_2$ are similar in both motion direction and location, the original position sequence representation is insufficient to differentiate them. This necessitates the development of more informative motion trajectory representation. However, most existing trajectory representation methods \cite{4,5,10,15} focus on performing transformation or parameterization on the original position sequence, while the problem of more informative representation
is not well addressed. Although some trajectory-modeling or local-modeling methods \cite{11,14,16,17,10} increase the informativeness of trajectories by including the contextual information among multiple trajectories, they only model partial contextual information from trajectories with similar patterns or trajectories in the same class. Thus, they still have limitations when differentiating ambiguous trajectories, such as trajectories near the boundary of similar trajectory classes (e.g., trajectories $A$ and $B$ in Fig.~\ref{fig:similar_example_a}).

\begin{figure*}
\centering
  \includegraphics[width=0.95\textwidth, height=4.09cm]{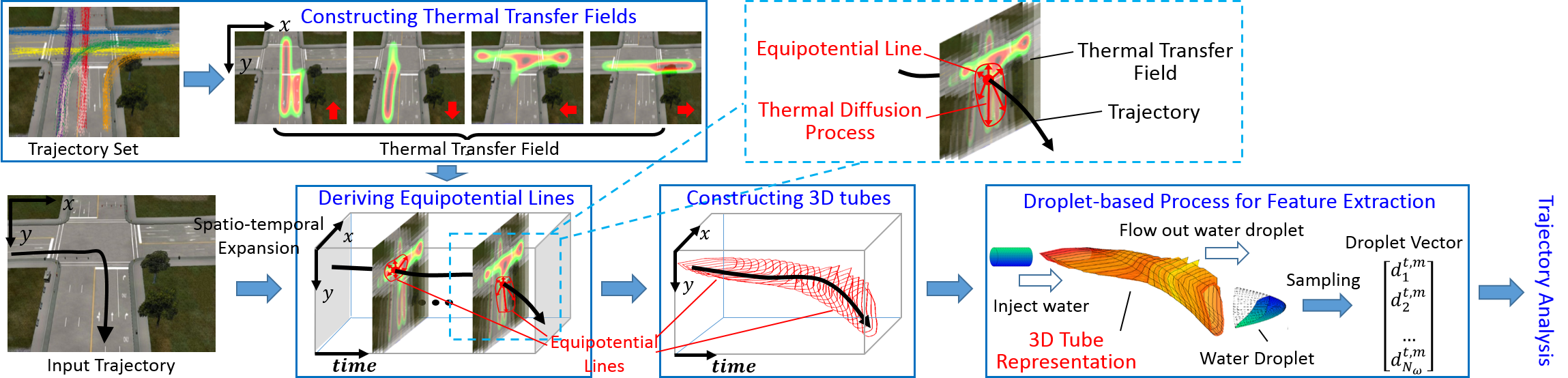}
  \vspace{-0.9mm}
  \caption{The framework of the proposed approach. It constructs a scene-specific thermal transfer field via trajectory training data; For a test trajectory sample, it builds a 3D tube based on the constructed thermal transfer field, and generates a feature vector via a droplet process. The obtained feature vector is applicable to various trajectory analysis applications. (Best viewed in color)}\label{fig:framework}
\end{figure*}

We argue that due to the stable constraint from a scene, the \emph{complete} contextual motion pattern around a trajectory provides an important cue for trajectory depiction. By \emph{complete}, we refer to the contextual motion information from \emph{all} given trajectories in the neighborhood of an input trajectory. For example, Fig.~\ref{fig:similar_example_b} shows the complete contextual motion patterns of two easily-confused trajectories $A$ and $B$ which are extracted from two similar trajectory classes $CP_1$ and $CP_2$. If we look at the contextual motion information in the neighborhood of trajectory $A$, there is a
strong left and slightly upward motion pattern (in green color) towards the end of $A$. In contrast, if we look at the contextual motion information in the neighborhood of trajectory $B$, there is an obvious upward motion pattern in the middle of $B$. Thus, the ambiguity between trajectories is expected to be reduced if the above-mentioned contextual motion information is properly modeled (cf. Fig.~\ref{fig:similar_example_c}).


\subsection{Our Work}

Based on this intuition, we develop a novel framework which utilizes the global motion trends in a scene to describe a trajectory. Specifically, for each point in a trajectory, we derive the dominant scene motion pattern around the point and utilize it to depict the point's complete contextual motion pattern. By integrating the dominant scene motion patterns for all points in a trajectory, we are able to describe a trajectory in a highly informative way and make fine distinctions among similar trajectories. The framework of our approach is shown in Fig.~\ref{fig:framework}.


Given a set of trajectories, thermal transfer fields are first constructed to describe the global motion pattern in the scene (cf. module `constructing thermal transfer fields' in Fig.~\ref{fig:framework}). Then, for an input trajectory, we expand it over the time domain to construct a 3D spatio-temporal curve, and derive an equipotential line for each point in this curve (cf. module `deriving equipotential lines' in Fig.~\ref{fig:framework}). The equipotential lines are decided by the locations of spatio-temporal curve points and their surrounding dominant scene motion patterns defined in thermal transfer fields. In this way, the complete contextual motion pattern can be captured.

After obtaining equipotential lines, a 3D tube is constructed which concatenates equipotential lines according to the temporal order of spatio-temporal curve points (cf. module `constructing 3D tubes' in Fig.~\ref{fig:framework}). This 3D tube is able to depict an input trajectory in a highly informative way, where the movement of a trajectory and the contextual motion pattern around a trajectory are effectively captured by the route and shape of the 3D tube.

Finally, a droplet-based process is applied which `injects' water in one end of a 3D tube and achieves a water droplet flowed out from the other end. This water droplet is further sampled to achieve a low-dimensional droplet vector to characterize the 3D tube shape (cf. module `droplet-based process for feature extraction' in Fig.~\ref{fig:framework}). Since different trajectories are depicted by 3D tubes with different shapes, by suitably modeling the water flow process, the droplet vector can precisely catch the unique characteristics of a 3D tube. The droplet vector will serve as the final trajectory representation format and is applied in trajectory analysis.


In summary, our contributions are three folds.
\squishlist
 \item We construct a thermal transfer field to describe the global motion pattern in a scene, derive equipotential lines to capture the contextual motion information of trajectory points, and introduce a 3D tube by concatenating equipotential lines to represent a motion trajectory. These components establish a novel framework for addressing the informative trajectory representation problem.
 \item Under this framework, we develop a droplet-based process which derives a simple but effective low-dimensional droplet vector to characterize the high-dimensional information in a 3D tube. The derived droplet vector can not only capture the characteristics of a 3D tube, but also suitably reduce the disturbance from trajectory noises.
 \item We investigate our tube-and-droplet representation to various trajectory analysis applications including trajectory clustering, trajectory classification \& abnormality detection, and 3D action recognition, and achieve the state-of-the-art performance.
\squishend


\section{Related Works\label{section:related_work}}

\subsection{Trajectory Representation and Modeling}

Properly representing motion trajectories is crucial to trajectory analysis. Many algorithms \cite{11,1,2,8,58} have been proposed for trajectory representation. Most of them \cite{1,4,5,15} aim to find suitable parameter sets to describe trajectories. Discrete Fourier transform (DFT) coefficients~\cite{5} and polynomial curve fitting~\cite{4} are two examples. Rao et al. \cite{1} extracted dynamic instants and used them as the key points to represent the spatio-temporal curvature of trajectories. These methods focus more on the effective representation of a trajectory's position sequence, while the issue of more \emph{informative} representation is not addressed. Although some works \cite{14,1,2,3,4} increase the informativeness of trajectory representation by introducing additional information such as motion velocity, temporal order, or object size, the added information is still restricted within a single trajectory and cannot be used to discriminate trajectories with similar patterns.

Trajectory-modeling methods that include the contextual information of multiple trajectories are also proposed \cite{11,17,10}. These methods construct probability models for each trajectory class and utilize them to guide the trajectory analysis process. Kim et al. \cite{11} introduced Gaussian process regression flows to model the location and velocity probability for each trajectory class. Morris and Trivedi \cite{17} clustered trajectories into spatial routes and encoded the spatio-temporal dynamics of each cluster by hidden Markov models (HMM). Hu et al. \cite{10} built a time-sensitive Dirichlet process mixture model (tDPMM) to represent the spatio-temporal characteristics of each trajectory class. These methods only focus on modeling the contextual information inside each individual class, while ignoring information from external trajectory classes. Therefore, they still have limitations when differentiating ambiguous trajectories, such as trajectories near the boundary of similar trajectory classes.

Another line to integrate contextual information is the local-modeling methods \cite{13,16}. These methods aim to utilize dynamic models or topic models to describe trajectories' local motion patterns. Nascimento et al. \cite{13} introduced low-level dynamic models to decompose trajectories into basic displacement patterns, and recognized trajectories based on the switch among these low-level models. Wang et al. \cite{16} proposed a non-parametric topic model called Dual Hierarchical Dirichlet Processes (Dual-HDP), which constructs local semantic regions including similar trajectory segments and performs trajectory analysis accordingly. Although these models share the contextual information from multiple trajectory classes, only the contextual information from similar trajectory segments is considered. Therefore, they are less effective when trajectories have large variations or reliable local models cannot be constructed due to insufficient similar trajectory segments.

The proposed 3D tube representation differs with existing approaches in the following aspects:
\squishlist
 \item We model the \emph{complete} contextual information around a trajectory, not only the contextual information from partial trajectories. This enables us to precisely catch the subtle changes among different trajectory classes.
 \item Most existing works handle trajectory analysis with complex probability models, which require sufficient trajectory data to construct reliable models. We establish a novel framework to model trajectories in a simple but effective way, which can work robustly under relatively small data size. Experimental evaluation shows we can achieve state-of-the-art performance on benchmark datasets with this simple procedure.
 \item Existing methods focus on the abstract modeling of trajectory information where the modeled trajectory information cannot be easily visualized. Our 3D tube representation is able to visualize a variety of trajectory information, including spatial-temporal movements, contextual motion patterns, and possible motion directions, in a clear and unified way.
\squishend

\subsection{Handling High-Dimensional Representations}

Since highly informative trajectory representation often leads to a complex and high-dimensional representation format, it is non-trivial to find suitable ways to handle this high-dimensional representation for trajectory analysis.

In \cite{18}, Euclidean distance and dynamic time wrapping (DTW) distance were utilized to measure the distance of two time-series trajectory sequences. Vlachos et al. \cite{33} further introduced the longest common subsequence (LCSS) distance. While these methods can handle the original position sequence format of a trajectory, they are not suitable to process higher dimensional representations.

In order to handle higher-dimensional data representations, some dimension reduction approaches were developed \cite{21,34}. However, due to the large variation and ambiguity among trajectories, trajectory representations often have complex distributions. Thus, simply applying dimension reduction cannot achieve satisfactory results. Moreover, some advanced manifold approaches were also introduced \cite{24,47,48}. In~\cite{24}, Lui modeled high-dimensional inputs as high-order tensors. The similarity between inputs were measured by the intrinsic distance between tensors, which is estimated through manifold charting. Although these methods can achieve better results when processing high-dimensional trajectory representations, they have considerably high computation complexity. Thus they are difficult to be applied on large scale trajectory analysis.

Besides dimension reduction, other works \cite{14,22,23} aim to develop proper distance metrics to measure the similarity between high-dimensional inputs. Lin et al. \cite{14} introduced a surface matching distance to measure the similarity between high-dimensional surface shapes. Sangineto \cite{22} and Gao et al. \cite{23} developed advanced Hausdorff distance which treats each high-dimensional input as a set of points and estimates the similarity between inputs from the distance between point sets. These methods are dependent on the quality of high-dimensional inputs, and noisy inputs will adversely affect the performance of these methods.

Different from the previous methods, we develop a novel droplet-based process which simulates the physical \emph{water flow} process and derives a low-dimensional droplet vector to characterize the high-dimensional 3D tube shape. This process has low complexity and can suitably reduce the disturbance from trajectory noises.

\section{3D Tube Representation \label{section:3D tube representation}}

In order to include the complete contextual motion information in trajectory representation, it is important to properly model and embed the global motion pattern information in a scene.
More specifically, assuming that there are $M$ trajectories available from a scene, denoted as $\{\mathbf{\Gamma}^m\}_{m=1}^{M}$.
Each trajectory is represented as $\mathbf{\Gamma}^m=\{ \mathbf{p}_n^m \}_{n=1}^{\mathcal{L}^m}$, where $\mathbf{p}_n^m\in \mathbb{R}^2$ is the \emph{position} of the $n$th point of the trajectory and ${\mathcal{L}^m}$ is the length of trajectory $\mathbf{\Gamma}^m$.
Accordingly, the \emph{speed} of the point in the position is calculated as $\mathbf{u}_n^m = \frac{\mathbf{p}_{n+1}^m-\mathbf{p}_n^m}{\Delta t}\in \mathbb{R}^2$, where $\Delta t$ is the sampling interval between adjacent points.
We aim to find the scene's global motion pattern which best describes the motion trends provided by $M$ given trajectories\footnote{Note that we do not differentiate given trajectories' class labels, and directly use \emph{all} given trajectories to find global motion patterns.}, and use this global motion pattern to derive the complete contextual motion patterns for positions along the route of an input trajectory. In this way, we are able to construct a 3D tube from these contextual motion patterns and obtain an informative representation for the input trajectory.

In this paper, we borrow the idea from thermal propagation~\cite{35,37,49} and introduce a trajectory model, which finds global motion pattern, derives contextual motion patterns, and constructs 3D tube representations with thermal propagation processes.

\subsection{Constructing Thermal Transfer Fields \label{section:fields}}

First, we model the global motion pattern in a scene based on the works in fluid thermal propagation \cite{35,37,49}. Specifically, the aggregation of $M$ given trajectories is modeled as a `\emph{fluid}' in the scene, where each $\mathbf{p}_n^m$ is a sample of the fluid and the corresponding $\mathbf{u}_n^m$ refers to the movement of the fluid which results in the transfer of thermal energies in position $\mathbf{p}_n^m$.

According to thermal transmission theories \cite{37,35}, the thermal diffusion result of the entire fluid is affected by a scene-related \emph{thermal transfer field} which decides the thermal propagation strengths at different positions and in different directions.
Therefore, by constructing an optimal thermal transfer field that best suits the thermal dynamics of the fluid defined by given trajectories, the fluid's thermal dynamics, which characterize the motion pattern from $M$ given trajectories, can be effectively embedded in the thermal transfer field.

In this work, we construct a scene's thermal transfer field based on the strategy of finite-element analysis \cite{49}.
Formally, we first segment the scene into grids of positions $\mathcal{G}=[1,...,W]\times[1,...,H]$, where $W$ and $H$ are the width and height of the scene.
Then, the thermal transfer field of the scene can be represented as:
\begin{equation}
\mathbf{K}=[k(\mathbf{p},\mathbf{a})]_{W\times H\times |\mathcal{A}|}
\label{equation:eqfield}
\end{equation}
where $k(\mathbf{p},\mathbf{a})$ ($\mathbf{p} \in \mathcal{G}$, $\mathbf{a} \in \mathcal{A}$) is the thermal transfer coefficient indicating the thermal propagation strength along direction $\mathbf{a}$ at position $\mathbf{p}$.
Here $\mathbf{a}$ is a normalized vector ($\|\mathbf{a}\|_2=1$) indicating thermal propagation directions, which is selected from a pre-defined direction set $\mathcal{A}$.
$|\mathcal{A}|$ counts the number of directions in the set. In this paper, $\mathcal{A}$ contains four directions, which depicts a scene's global motion pattern in upward ($y^-$), downward ($y^+$), leftward ($x^-$), and rightward ($x^+$) directions, respectively (cf. module `constructing thermal transfer fields' in Fig.~\ref{fig:framework}).

Assuming that the given trajectories correspond to a stable fluid, we can construct an optimal thermal transfer field $\mathbf{K}$ by minimizing the total amount of thermal energies being transferred during the fluid flow process, as:
\begin{align}
&\mathop{\min}_{\mathbf{K}} \sum_{\mathbf{a}\in\mathcal{A}}\sum_{\mathbf{p}\in\mathcal{G}}\Delta E(k(\mathbf{p},\mathbf{a}))\label{equation:eq3}\\
s.t.~~ & k(\mathbf{p},\mathbf{a})\ge0, \sum_{\mathbf{p}\in\mathcal{G}}k(\mathbf{p},\mathbf{a})=\kappa,~\mbox{for}~\mathbf{a}\in\mathcal{A}.\nonumber
\end{align}
where $\Delta E(k(\mathbf{p},\mathbf{a}))$ is the amount of thermal energy transported by the fluid from position $\mathbf{p}$ along direction $\mathbf{a}$ within a unit time
interval. $\kappa$ ($\kappa>0$) is a constant. The first constraint in (\ref{equation:eq3}) guarantees to achieve physically-meaningful thermal transfer fields (i.e., avoid negative transfer fields), and the second constraint guarantees to obtain proper distributions of $k(\mathbf{p},\mathbf{a})$ (i.e., avoid transfer fields becoming all infinite values). $\Delta E(k(\mathbf{p},\mathbf{a}))$ can be calculated from thermal transmission theories \cite{35,37}:
\begin{equation}
\Delta E(k(\mathbf{p},\mathbf{a}))=\eta\frac{\rho(\mathbf{p})u(\mathbf{p},\mathbf{a})}{k(\mathbf{p},\mathbf{a})}
\label{equation:eqpartial}
\end{equation}
Here $\eta$ is a parameter related to the temperature difference between a position $\mathbf{p}$ and its neighbors~\cite{35,37}. In our paper, since we want to focus on the relationship between $\Delta E$ and $k(\mathbf{p},\mathbf{a})$, we simply assume the temperature difference condition to be the same when calculating $\Delta E$ at different positions, and set $\eta$ as a constant. $\rho(\mathbf{p})$ is the \emph{density} of fluid at position $\mathbf{p}$, $u(\mathbf{p},\mathbf{a})$ is the moving \emph{velocity} along direction $\mathbf{a}$ at position $\mathbf{p}$. Physically, $\rho(\mathbf{p})u(\mathbf{p},\mathbf{a})$ measures the number of fluid particles passing through position $\mathbf{p}$ along direction $\mathbf{a}$ within a unit time interval.
${k(\mathbf{p},\mathbf{a})}$ indicates the efficiency of thermal energy transfer along direction $\mathbf{a}$ at $\mathbf{p}$. As a result,
${k(\mathbf{p},\mathbf{a})}\Delta E$ refers to the amount of thermal energies actually received by $\mathbf{p}$'s neighboring position when an amount of energy $\Delta E$ is transferred out from $\mathbf{p}$~\cite{35,37}.

According to (\ref{equation:eq3}) and (\ref{equation:eqpartial}), we let a fluid flow along the aggregated routes of $M$ given trajectories, and measure the total amount of thermal energy transfers over all positions. The thermal transfer field that leads to the smallest total transferred energy will be the optimal field that best suits the scene.

More specifically, from (\ref{equation:eqpartial}), the amount of energy transfer $\Delta E$ is jointly decided by $k(\mathbf{p},\mathbf{a})$ (thermal transfer coefficient), $\rho(\mathbf{p})$ (fluid's density), and $u(\mathbf{p},\mathbf{a})$ (fluid's velocity). When $\rho(\mathbf{p})u(\mathbf{p},\mathbf{a})$ increases, the amount of fluid flowing along direction $\mathbf{a}$ at $\mathbf{p}$ becomes stronger, which leads to larger chances of energy transfer. Thus, by minimizing $\sum_{\mathbf{a,p}}\Delta E$ in (\ref{equation:eq3}), ${k(\mathbf{p},\mathbf{a})}$ is proportionally adjusted with $\rho(\mathbf{p})u(\mathbf{p},\mathbf{a})$ such that ${k(\mathbf{p},\mathbf{a})}$ with higher energy transfer efficiencies are assigned to positions/directions with stronger fluid flows. In this way, the resulting thermal transfer field can properly suit the thermal dynamics of the fluid.

$\rho(\mathbf{p})$ in (\ref{equation:eqpartial}) is calculated based on the nonparametric estimation of sample points $\mathbf{p}^m_n$ in given trajectories:
\begin{equation}
\rho(\mathbf{p})=\sum_{m=1}^{M}\sum^{{\mathcal{L}^m}}_{n=1} \exp\left({-\frac{\lVert\mathbf{p}-\mathbf{p}^m_n\rVert}{2\sigma^2}}\right).\label{equation:rho}
\end{equation}
Similarly, $u(\mathbf{p},\mathbf{a})$ in (\ref{equation:eqpartial}) is the nonparametric estimation of sample points and their corresponding projected speeds:
\begin{equation}
u(\mathbf{p},\mathbf{a}) = \sum_{m=1}^{M}\sum^{{\mathcal{L}^m}}_{n=1} \max(\mathbf{a}^{\top}\mathbf{u}_n^m,0) \exp\left({-\frac{\lVert\mathbf{p}-\mathbf{p}^m_n\rVert}{2\sigma^2}}\right)
\label{equation:eq4}
\end{equation}
where $\mathbf{a}^{\top}\mathbf{u}_n^m$ is the projection of speed $\mathbf{u}_n^m$ in $\mathbf{a}$-th direction. $\max(\mathbf{a}^{\top}\mathbf{u}_n^m,0)$ ensures that only the directions with positive velocity are considered.

(\ref{equation:rho}) and (\ref{equation:eq4}) ensure that $\rho(\mathbf{p})u(\mathbf{p},\mathbf{a})$ can correctly reflect the global motion pattern contained in given trajectories. For example, when a large number of trajectories pass through a position $\mathbf{p}$ with relatively high speed, $\rho(\mathbf{p})u(\mathbf{p},\mathbf{a})$ will become a large number. On the contrary, for positions where trajectories rarely pass, $\rho(\mathbf{p})u(\mathbf{p},\mathbf{a})$ will be small.
Moreover, note that $\rho(\mathbf{p})u(\mathbf{p},\mathbf{a})$ in (\ref{equation:rho}) and (\ref{equation:eq4}) is estimated by integrating the information of \emph{all} trajectories without differentiating their classes. Therefore, the resulting thermal transfer fields in fact include the \emph{complete} contextual motion information among given trajectories. This enables the embedding of the complete contextual motion pattern in later 3D tube construction and trajectory representation steps.

Based on (\ref{equation:eqpartial})-(\ref{equation:eq4}), (\ref{equation:eq3}) can be solved by the \emph{Cauchy--Schwarz inequality} and the optimal solutions turns out to be:
\begin{equation}
k(\mathbf{p},\mathbf{a})\propto\sqrt{\rho(\mathbf{p})u(\mathbf{p},\mathbf{a})} \qquad
\label{equation:eq5}
\end{equation}

Figs~\ref{fig:framework} and~\ref{fig:3D_tube_field} show some examples of the constructed thermal transfer fields from given trajectories. We can see that the thermal transfer
fields constructed by our approach can effectively capture
the four-direction global motion patterns contained in given
trajectories.
For example, in Fig.~\ref{fig:rabc}, both the upward and rightward motion patterns of the blue trajectories in~\subref{fig:3D_tube_traj} are effectively embedded in the \emph{upward} and \emph{rightward} thermal transfer fields in~\subref{fig:3D_tube_field}.

\subsection{Deriving Equipotential Lines \label{section:equilines}}

\begin{figure}
  \centering
  \subfloat[]{\includegraphics[height=1.57 cm, width= 1.85 cm]{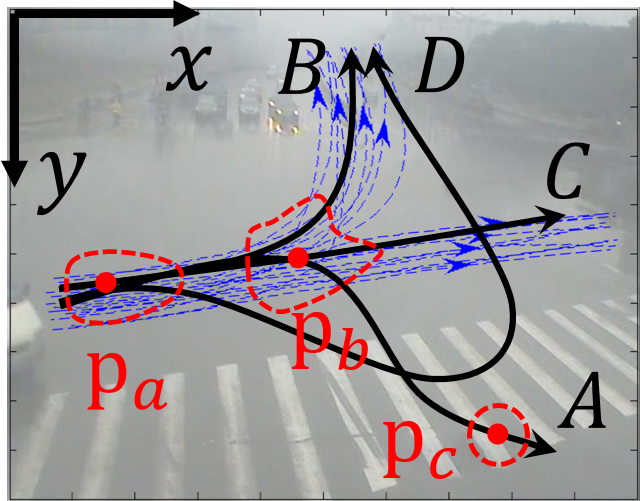}\label{fig:3D_tube_traj}}
  \hspace{1mm}
  \subfloat[]{\includegraphics[height=1.57 cm, width=6.8 cm]{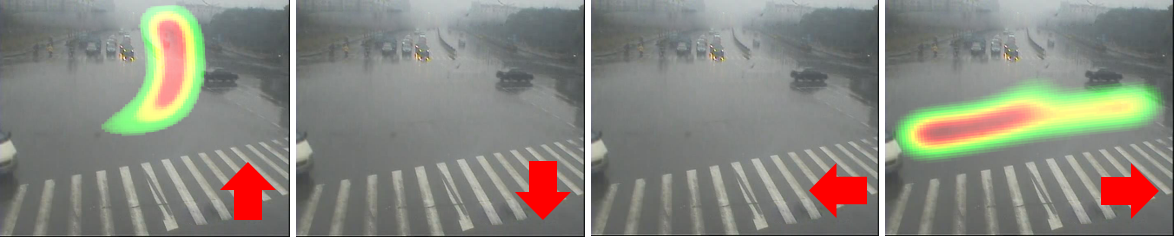}\label{fig:3D_tube_field}} \\
  \caption{(a) Input trajectories (black solid curves) and given trajectories (blue dashed curves). (b) Thermal transfer fields constructed from the blue dashed curves in (a). Figures from left to right correspond to thermal transfer fields in directions upward ($y^-$), downward ($y^+$), leftward ($x^-$), rightward ($x^+$), respectively. (Best viewed in color)}\label{fig:rabc}
\end{figure}

Based on the constructed thermal transfer field $\mathbf{K}$, we can perform thermal propagation for each point in an input trajectory, so as to embed the input trajectory's contextual motion information. More specifically, we first obtain a \emph{thermal diffusion map} for an input trajectory point $\mathbf{p}_n^m$, which depicts the dominant scene motion pattern around $\mathbf{p}_n^m$ by propagating thermal energies from $\mathbf{p}_n^m$ to the entire scene through $\mathbf{K}$. Then we derive a constant-energy line called \emph{equipotential line} from this map, and use it to capture the contextual motion information for the trajectory point.

Formally, for each trajectory point $\mathbf{p}_n^m$, we denote its corresponding thermal diffusion map as $\mathbf{E}(\mathbf{p}_n^m)=[E(\mathbf{p}_n^m, \mathbf{p})]_{W\times H}$, $\mathbf{p}\in\mathcal{G}$, where $E(\mathbf{p}_n^m,\mathbf{p})$ is the thermal energy at position $\mathbf{p}$ in $\mathbf{p}_n^m$'s map.
Taking $\mathbf{E}(\mathbf{p}_n^m)$ as a function of time $t$ and direction $\mathbf{a}\in\mathcal{A}$, we calculate $\mathbf{E}(\mathbf{p}_n^m)$ via an iterative method.
Specifically, the initial map $\mathbf{E}_0(\mathbf{p}_n^m)$ is defined as follows:
\begin{equation}
        E_0(\mathbf{p}_n^m,\mathbf{p})=
        \left\{
           \begin{aligned}
           &E_\epsilon,~~~~~\text{if}~~\mathbf{p}=\mathbf{p}_n^m\\
           &0,~~~~~~~\text{otherwise}
           \end{aligned}
        \right. \,,   \label{equation:init}
\end{equation}
where $E_0(\mathbf{p}_n^m,\mathbf{p})$ is the initialized thermal energy at position $\mathbf{p}$.
$E_\epsilon=100$ is a constant.
According to (\ref{equation:init}), the thermal diffusion map of trajectory point $\mathbf{p}_n^m$ is initialized by assigning a large thermal energy to $\mathbf{p}_n^m$'s position and assigning zero energies to other positions.
Then, a thermal diffusion process is applied which transfers thermal energies from $\mathbf{p}_n^m$ to other positions through the constructed thermal transfer field, and creates an energy-propagated thermal diffusion map.
The thermal diffusion process is defined by \cite{35}:
\begin{equation}
\label{equation:eqthermal}
\frac{\partial E(\mathbf{p}_n^m,\mathbf{p})}{\partial t}= \sum_{\mathbf{a}\in\mathcal{A}}k(\mathbf{p},\mathbf{a})\frac {\partial^2 E(\mathbf{p}_n^m,\mathbf{p})} {\partial {\mathbf{a}}^2}.
\end{equation}
Here $\frac{\partial^2 E(\mathbf{p}_n^m,\mathbf{p})}{\partial {\mathbf{a}}^2}$ means the 2nd order partial derivative of $E(\mathbf{p}_n^m,\mathbf{p})$ along direction $\mathbf{a}$.
$k(\mathbf{p},\mathbf{a})$ is a thermal transfer coefficient in the constructed thermal transfer field $\mathbf{K}$ (cf. (\ref{equation:eqfield})-(\ref{equation:eq3})).
From (\ref{equation:eqthermal}), the thermal diffusion process is mainly controlled by the thermal energy difference among neighboring positions and the thermal transfer coefficients from thermal transfer field.
Thus, by properly setting the initial energy difference (cf. (\ref{equation:init})), the resulting thermal diffusion map can effectively capture the desired contextual motion pattern contained in the thermal transfer field.

\begin{figure}
  \centering
  \subfloat[]{\includegraphics[height=2.35cm, width=3.3cm]{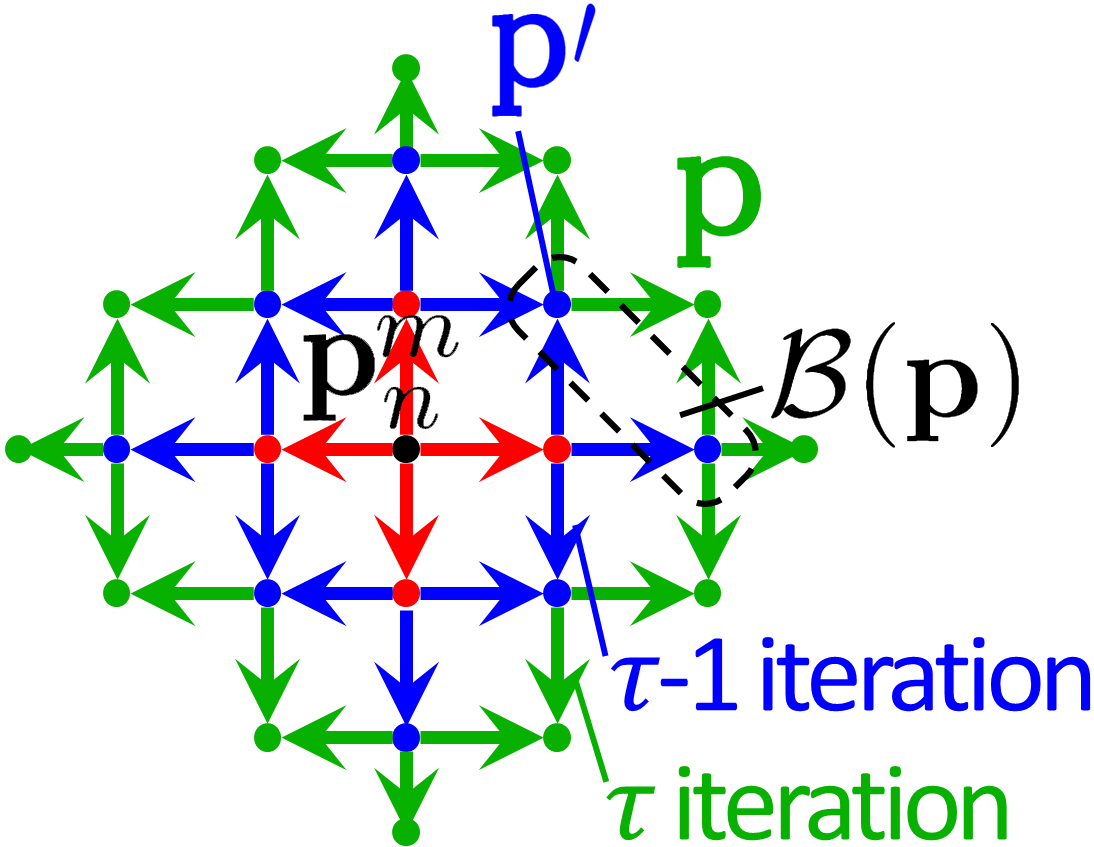}\label{fig:rabc_field}}
  \hspace{3mm}
  \subfloat[]{\includegraphics[height=2.4 cm, width=1.3 cm]{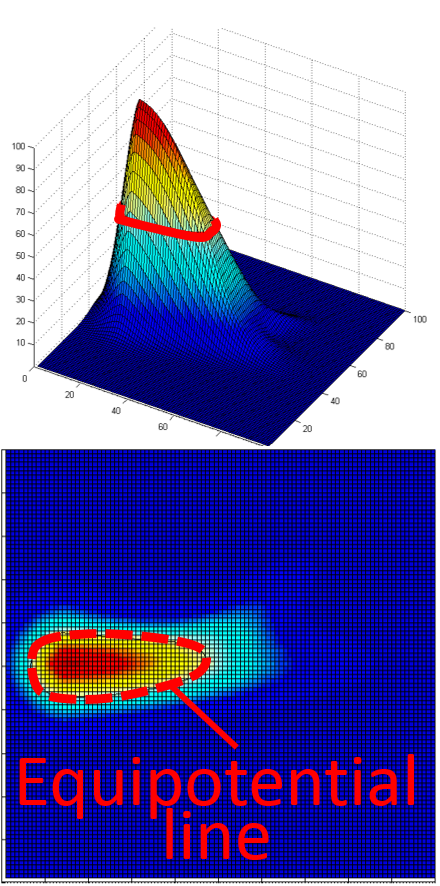}\label{fig:rabc_a}}
  \hspace{3mm}
  \subfloat[]{\includegraphics[height=2.4 cm, width=1.3 cm]{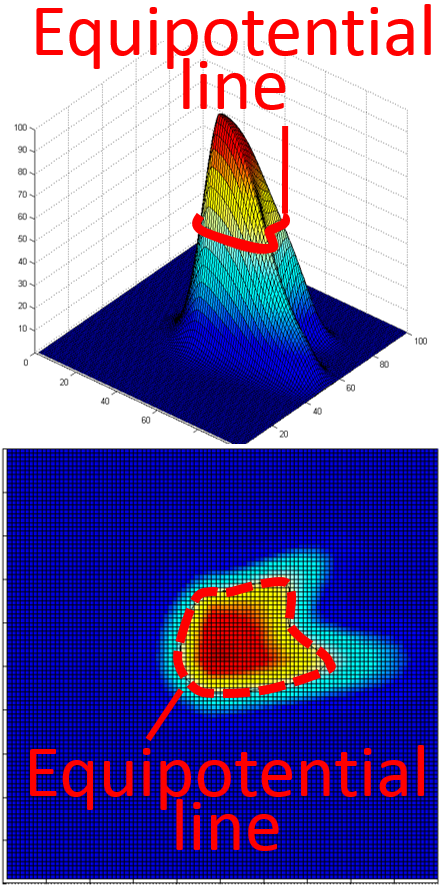}\label{fig:rabc_b}}
  \hspace{3mm}
  \subfloat[]{\includegraphics[height=2.4 cm, width=1.3 cm]{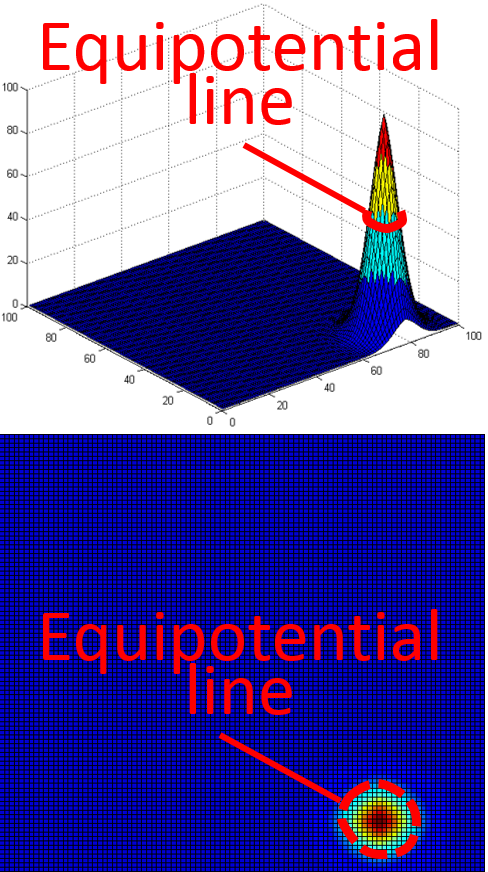}\label{fig:rabc_c}}
  \caption{Examples of equipotential lines. (a) The process of iterative thermal diffusion. (b)-(d) Lower figures: The thermal diffusion maps and the equipotential lines for trajectory points $\mathbf{p}_a$, $\mathbf{p}_b$, and $\mathbf{p}_c$ in Fig.~\ref{fig:3D_tube_traj}; Upper figures: The thermal diffusion maps and the equipotential lines displayed in 3D. (Best viewed in color)}\label{fig:rabc1}
\end{figure}

(\ref{equation:eqthermal}) is difficult to solve because: 1) The thermal diffusion process is performed by integrating the information from multiple directions $\mathbf{a}$; 2) The thermal transfer fields are non-homogeneous whose thermal transfer coefficients $k(\mathbf{p},\mathbf{a})$ vary over different locations. We propose an approximation solution which obtains thermal diffusion map in an iterative way, as illustrated in Fig.~\ref{fig:rabc_field}.

From Fig.~\ref{fig:rabc_field}, after initializing a thermal diffusion map by (\ref{equation:init}), the thermal energy at $\mathbf{p}_n^m$ is diffused iteratively outward to other positions.
During each iteration $\tau$, positions which were diffused in the previous iteration $\tau-1$ will diffuse energies to their outside neighboring positions (as indicated by arrows in Fig.~\ref{fig:rabc_field}).
According to the thermal diffusion process in (\ref{equation:eqthermal}), thermal diffusion between neighboring positions in one iteration can be approximated by:

\begin{equation}
E_{\tau}(\mathbf{p}_n^m, \mathbf{p})\approx
\frac{\sum_{\mathbf{p}'\in \mathcal{B}(\mathbf{p})}E_{\tau-1}(\mathbf{p}_n^m,\mathbf{p}')\exp\left(-\frac{\| \mathbf{p}-\mathbf{p}'\|}{k(\mathbf{p},\frac{\mathbf{p}-\mathbf{p}'}{\|\mathbf{p}-\mathbf{p}'\|})}\right)}{|\mathcal{B}(\mathbf{p})|}
\label{equation:eq7}
\end{equation}
where $\mathcal{B}(\mathbf{p})$ is the neighborhood of $\mathbf{p}$ and $|\mathcal{B}(\mathbf{p})|$ is the size of the neighborhood.
For $\mathbf{p}$ and its neighbors $\mathbf{p}'\in\mathcal{B}(\mathbf{p})$, $\lVert \mathbf{p}-\mathbf{p}'\rVert$ is their distance and $k(\mathbf{p},\frac{\mathbf{p}-\mathbf{p}'}{\|\mathbf{p}-\mathbf{p}'\|})$ is the thermal transfer coefficient along direction $\frac{\mathbf{p}-\mathbf{p}'}{\|\mathbf{p}-\mathbf{p}'\|}$ at position $\mathbf{p}$.

(\ref{equation:eq7}) reveals that: 1) The diffused thermal energy in a position $\mathbf{p}'$ is summed over all energies propagated from its neighboring positions in both vertical and horizontal directions. In this way, the complete contextual information can be included. 2) The thermal transfer coefficients $k(\mathbf{p},\mathbf{a})$ control the thermal diffusion results. In this way, the motion pattern information from the thermal transfer field can be properly reflected in the resulting thermal diffusion map. 3) The thermal diffusion map $\mathbf{E}(\mathbf{p}_n^m)$ for position $\mathbf{p}_n^m$ is fully decided by the thermal transfer field $\mathbf{K}$ (cf. (\ref{equation:init}) and (\ref{equation:eq7})). This implies that the contextual motion information  $\mathbf{E}(\mathbf{p}_n^m)$ for each position in a scene is fixed after a scene's global motion pattern $\mathbf{K}$ is determined. Therefore, if two input trajectories pass through position $\mathbf{p}_n^m$ with different routes, they will have the same thermal diffusion map at $\mathbf{p}_n^m$.

Fig.~\ref{fig:rabc1} shows some examples of thermal diffusion map results derived from the thermal transfer fields in Fig.~\ref{fig:3D_tube_field}. In Fig.~\ref{fig:rabc1}, \subref{fig:rabc_a}-\subref{fig:rabc_c} show the thermal diffusion maps of three points $\mathbf{p}_a$, $\mathbf{p}_b$, $\mathbf{p}_c$ on the black trajectory $A$ in Fig.~\ref{fig:3D_tube_traj}.

According to Fig.~\ref{fig:3D_tube_traj}, since moving rightward from $\mathbf{p}_a$ appears frequently in the scene (as there are lots of dashed blue trajectories moving rightward around $\mathbf{p}_a$), large \emph{rightward}-direction thermal transfer coefficients are obtained around $\mathbf{p}_a$. This allows more thermal energies being propagated to $\mathbf{p}_a$'s right side, thus leading to a long rightward tail in the thermal diffusion map of $\mathbf{p}_a$ (cf. Fig.~\ref{fig:rabc_a}). Similarly, since $\mathbf{p}_b$ is located in a region including frequent movements in both rightward and upright-ward directions, $\mathbf{p}_b$'s thermal diffusion map includes big tails in both directions and displays a $V$-like shape (cf. the lower figure in Fig.~\ref{fig:rabc_b}). Comparatively, since $\mathbf{p}_c$ is located in a region where seldom trajectories pass, the constructed thermal transfer coefficients are small in all directions around $\mathbf{p}_c$. This makes $\mathbf{p}_c$'s thermal diffusion map decay quickly around $\mathbf{p}_c$, as in Fig.~\ref{fig:rabc_c}. From the example of Figs~\ref{fig:rabc_a}-\ref{fig:rabc_c}, our thermal diffusion map indeed provides a reliable way to capture the complete and unique contextual motion patterns for input trajectory points.

In order to represent thermal diffusion maps in a more effective way, we further derive equipotential lines to capture the fundamental information of thermal diffusion maps. An equipotential line can be easily achieved by finding a constant-energy line on a thermal diffusion map. In this paper, we acquire constant-energy line whose energy decreases to half of the initial energy $E_\epsilon$, as indicated by the red circles in Figs~\ref{fig:rabc_a}-\ref{fig:rabc_c}.

\subsection{Constructing 3D Tubes \label{section:3Dtube}}

After deriving equipotential lines for all points in a trajectory, a 3D tube can be constructed to represent this trajectory by concatenating these equipotential lines according to their temporal order in the trajectory.

\begin{figure}
  \centering
  \subfloat[]{\includegraphics[width=0.11\textwidth]{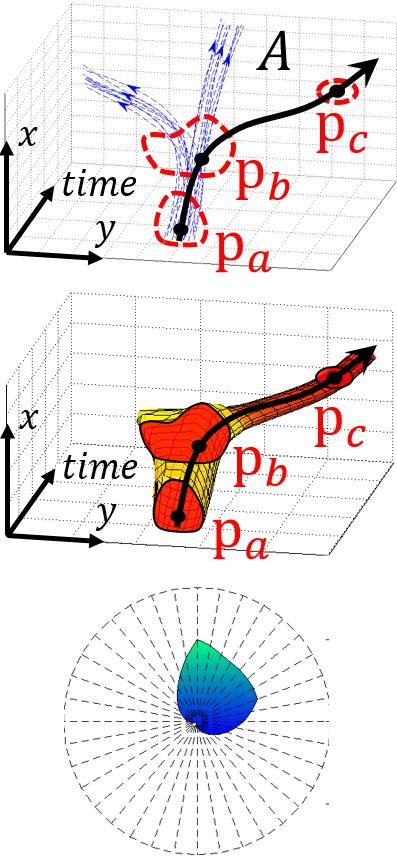}\label{fig:3D_tube_c}}
  \hspace{2mm}
  \subfloat[]{\includegraphics[width=0.11\textwidth]{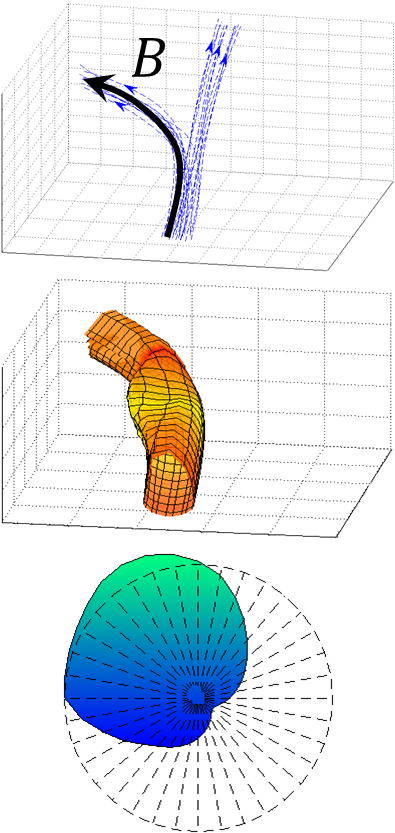}\label{fig:3D_tube_b}}
  \hspace{2mm}
  \subfloat[]{\includegraphics[width=0.11\textwidth]{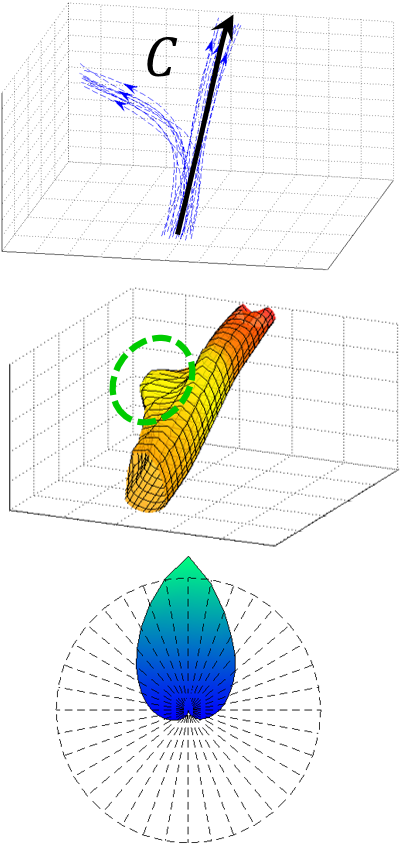}\label{fig:3D_tube_a}}
  \hspace{2mm}
  \subfloat[]{\includegraphics[width=0.11\textwidth]{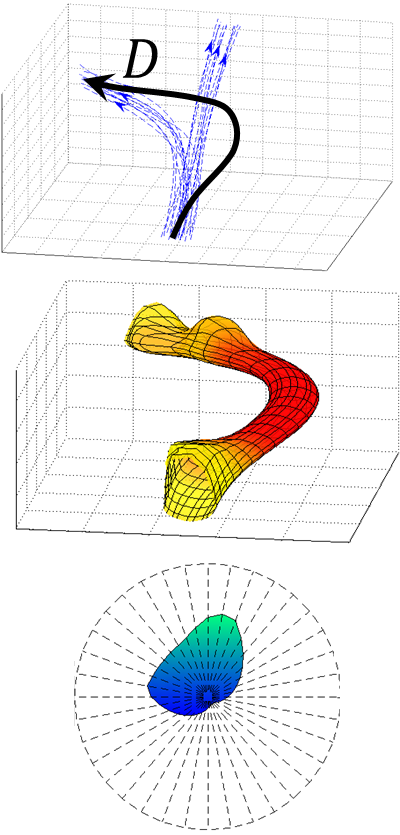}\label{fig:3D_tube_d}}
  \caption{Examples of 3D tubes and water droplet results. First row: results by expanding trajectories in Fig.~\ref{fig:3D_tube_traj} into 3D spatio-temporal curves; Second row: 3D tube representations for the black input trajectories $A$-$D$ in Fig.~\ref{fig:3D_tube_traj}; Third row: water droplet results derived from 3D tubes. (Note: In the middle row of (a)-(d), the thickness of a tube is represented by different colors where yellow indicates thick and red indicates narrow. Best viewed in color)}\label{fig:3D_tube}
\end{figure}

Fig.~\ref{fig:3D_tube} shows some 3D tube examples for the black input trajectories $A$-$D$ in Fig.~\ref{fig:3D_tube_traj}. The first row in Fig.~\ref{fig:3D_tube} shows the results by expanding trajectories in \ref{fig:3D_tube_traj} into 3D spatio-temporal curves. The second row of Fig.~\ref{fig:3D_tube} illustrates the 3D tube representations for trajectories $A$-$D$ in~\ref{fig:3D_tube_traj}. Furthermore, the equipotential lines for three points on one input trajectory (i.e., trajectory $A$ in Fig.~\ref{fig:3D_tube_traj}) are also displayed by red slices in Fig.~\ref{fig:3D_tube_c}. These red slices clearly show that a 3D tube is constructed by sequentially concatenating a trajectory's equipotential lines in a 3D spatio-temporal space.

From Fig.~\ref{fig:3D_tube}, we can observe that:
\squishlist
 \item The constructed 3D tube contains rich information about a trajectory, where both the movement and the contextual motion pattern are effectively embedded. For example, the route of a 3D tube represents the movement. The thickness variation of a 3D tube indicates whether there are frequent motion patterns in the context around a trajectory (e.g., a 3D tube will become narrow if a trajectory goes through a region where trajectories rarely pass, such as trajectory $A$ in Fig.~\ref{fig:3D_tube_c}). Moreover, the shapes of equipotential lines in a 3D tube also indicate possible motion trends provided by the contextual motion patterns. For example, the convex part circled by the green dashed line in the second row of Fig.~\ref{fig:3D_tube_a} indicates that moving upleft-ward around $\mathbf{p}_b$ (i.e., turn left in the original 2D scene) is another possible motion trend which also appears frequently in the scene. More discussions about the informativeness of 3D tube representation will be provided in the experimental results.
\item Different from the previous trajectory modeling methods \cite{10,17} whose modeled trajectory information cannot be easily visualized, our 3D tube representation is able to visualize information of a trajectory in a clear and unified way. For example, one can easily observe a trajectory's movement and contextual motion pattern from the route and shape variation of its 3D tube representation. This in fact provides a useful tool for people to intuitively observe and analyze trajectory information.
\squishend
\section{The Water-droplet Process \label{section:droplet method}}

After constructing 3D tubes for input trajectories, we need to find suitable ways to effectively handle the high-dimensional information included in 3D tubes. In this paper, we introduce a droplet-based process which simulates the physical water flow process \cite{37} and derives a low-dimensional droplet vector to characterize a high-dimensional 3D tube shape.

\begin{figure}
  \centering
  \subfloat[]{\includegraphics[width=0.4\textwidth, height=0.195\textwidth]{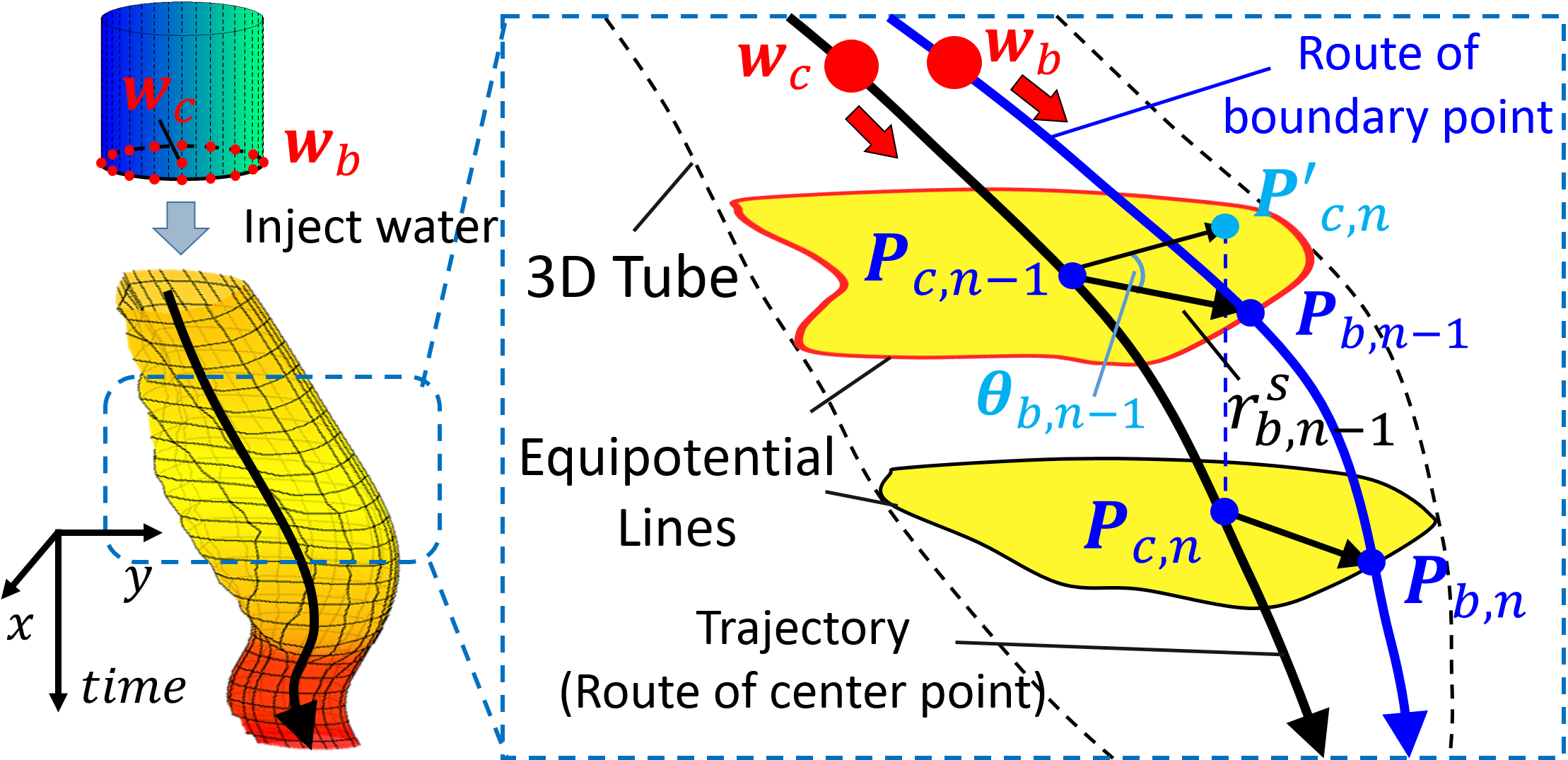}\label{fig:water_b}} \\
  \vspace{-3mm}
  \subfloat[]{\includegraphics[width=0.34\textwidth, height=0.102\textwidth]{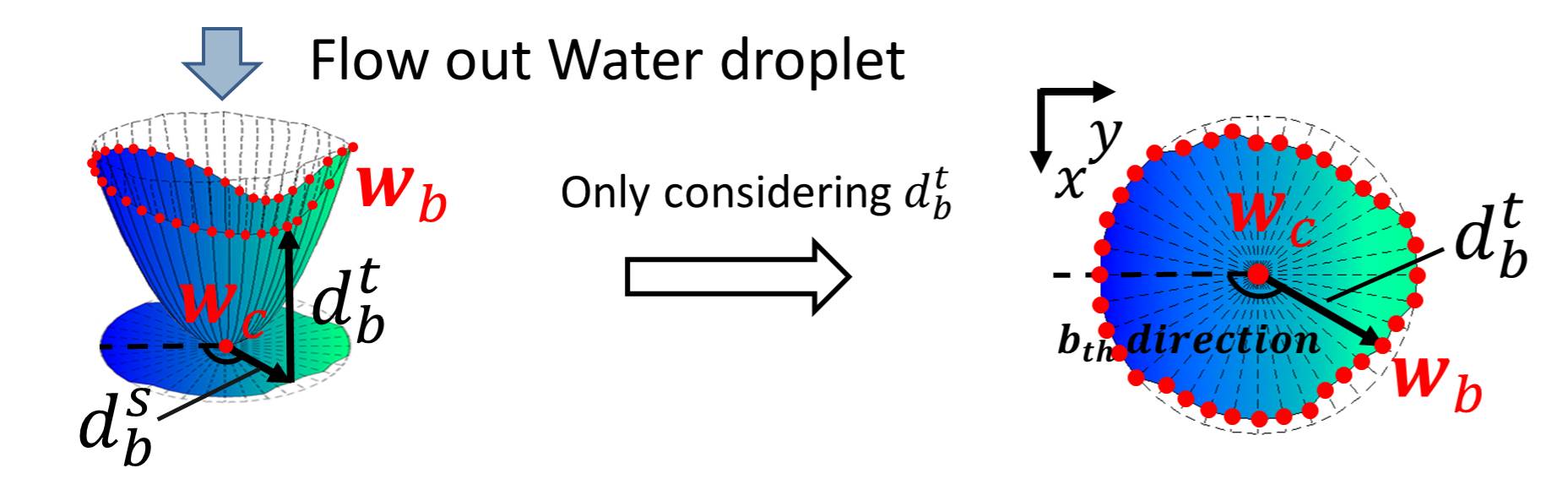}\label{fig:water_a}}
  \vspace{-2mm}
  \caption{The water droplet process. (a) Illustrations of the input water droplet, $\mathbf{P}_{b,n}$, $\mathbf{P}_{c,n}$, $\theta_{b,n}$, and routes of water droplet points $\mathbf{w}_c$, $\mathbf{w}_b$ when passing through a 3D tube. (b) The resulting 3D water droplet (left) and the simplified 2D droplet by only considering $d_{b}^t$ (right).}\label{fig:water_droplet}
\end{figure}

The process of the droplet-based process is displayed in Fig.~\ref{fig:water_droplet}. We inject a drop of water with fixed shape in one end of a 3D tube (cf. Fig.~\ref{fig:water_b}) and achieve a water droplet flowed out from the other end (cf. Fig.~\ref{fig:water_a}, note that the water is flowed along the time-axis in a 3D spatial-temporal space). According to fluid mechanic theories \cite{37}, when a water droplet passes through a 3D tube, its shape will be affected by the friction and extrusion forces from the boundary of the tube. 3D tubes with different shapes will provide different impacts to water passing through them, and create different droplet outputs. Therefore, by properly designing the water flow process, the derived droplet can effectively capture the characteristics of a high-dimensional 3D tube shape.

A water droplet is described by a center point $\mathbf{w}_c$ and a set of boundary points $\{\mathbf{w}_{b}\}_{b=1,\dots,N_w}$ where $N_w=36$ is the total number of boundary points being considered. Moreover, two distances are defined to represent the relative location between a boundary point $\mathbf{w}_{b}$ and $\mathbf{w}_c$: 1) $d_{b}^t$ is the distance between $\mathbf{w}_{b}$ and $\mathbf{w}_c$ in the time axis; 2) $d_{b}^s$ is the distance between $\mathbf{w}_{b}$ and $\mathbf{w}_c$ in the spatial plane, as in Fig.~\ref{fig:water_a}. In order to simplify calculation, we only calculate $d_{b}^t$ at the output of a 3D tube, such that a 3D droplet can be simply described by $d_{b}^t$ in a 2D plane, as in Fig.~\ref{fig:water_a}. Therefore, in the following, we focus on discussing $d_{b}^t$.

Before water starts to flow in a 3D tube, $d_{b}^t$ are initialized as $0$ to make all boundary points located on the same spatial plane as the center point $\mathbf{w}_c$, as in the upper figure in Fig.~\ref{fig:water_b}. Then, during the water flow process, we let the center point $\mathbf{w}_c$ follow the route of a 3D tube's input trajectory (i.e., the input trajectory that creates the 3D tube), as indicated by ${\{\mathbf{P}_{c,n}\}}^{\mathcal{L}}_{n=1}$ , where $\mathcal{L}$ is the length of the input trajectory and $\mathbf{P}_{c,n}$ is the location of the $n$th input-trajectory point in the spatial-temporal space (cf. the black line in Fig.~\ref{fig:water_b}). Similarly, a boundary point $\mathbf{w}_{b}~(b\in\{1,\dots,N_w\})$ will follow the route constructed by the $b$th boundary points on a 3D tube:  ${\{\mathbf{P}_{b,n}\}}^{\mathcal{L}}_{n=1}$, where $\mathbf{P}_{b,n}$ is the position located on $b$th direction of $\mathbf{P}_{c,n}$ on $\mathbf{P}_{c,n}$'s equipotential line (cf. the blue line in Fig.~\ref{fig:water_b}).

When the center point $\mathbf{w}_c$ of a water droplet passes through a 3D tube with a constant velocity in the time-axis, the velocity of the droplet's boundary points $\mathbf{w}_{b}~(b=1,\dots,N_w)$ will be jointly affected by two forces: the viscosity force from $\mathbf{w}_c$ which pulls $\mathbf{w}_{b}$ close to $\mathbf{w}_c$, and the friction force from a 3D tube's boundary which resists $\mathbf{w}_{b}$ from approaching $\mathbf{w}_c$ \cite{37}. Since both forces are controlled by the shape of a 3D tube, by calculating the relative distances $d_{b}^t$ between $\mathbf{w}_c$ and $\mathbf{w}_{b}$ at a tube's output, the characteristics of a 3D tube can be properly captured.

The time-axis distance $d_{b}^t$ between points $\mathbf{w}_c$ and $\mathbf{w}_{b}$ at the output of a tube (cf. Fig.~\ref{fig:water_a}) can be approximated by:
\begin{equation}
d_{b}^t\propto\frac{1}{\mathcal{L}}\sum^\mathcal{L}_{n=1}{(v_{c,n}-v_{b,n})} \label{equation:eq8}
\end{equation}
where $\mathcal{L}$ is the length of a 3D tube. $v_{c,n}$ and $v_{b,n}$ are the time-axis velocities when points $\mathbf{w}_c$ and $\mathbf{w}_b$ pass through positions $\mathbf{P}_{c,n}$ and $\mathbf{P}_{b,n}$, respectively. $v_{c,n}=v_\mathcal{C}$ meaning that $\mathbf{w}_c$ passes through different $\mathbf{P}_{c,n}$ with a constant time-axis velocity $v_\mathcal{C}$. $\frac{1}{\mathcal{L}}$ is included to normalize $d_{b}^t$.

From (\ref{equation:eq8}), the time-axis distance $d_{b}^t$ is measured by accumulating the velocity difference $v_{c,n}-v_{b,n}$ between points $\mathbf{w}_c$ and $\mathbf{w}_{b}$ when passing through a 3D tube. By performing velocity accumulation, we are able to characterize high-dimensional tube information in a relative simple way while reducing the disturbance from trajectory noises.

According to fluid viscosity theories~\cite{26}, $v_{c,n}-v_{b,n}$ in (\ref{equation:eq8}) can be approximated by (see Appendix in the supplementary material for details):
\begin{equation}
v_{c,n}-v_{b,n}\approx\left(1-\frac{\lambda_1}{r_{b,n-1}^s}\right)(v_\mathcal{C}-v_{b,n-1})+\lambda_2 \cos\theta_{b,n-1} \label{equation:eqvelocity}
\end{equation}
where $\lambda_1$ and $\lambda_2$ are constant coefficients. $v_{b,n}$ is the time-axis velocity of $\mathbf{w}_b$ when passing through position $\mathbf{P}_{b,n}$. $\mathbf{w}_b$'s velocity is initialized as $0$ before the water flow process (i.e., $v_{b,0}=0$). $r_{{b},n-1}^s$ is the distance from ($n-1)-$th trajectory position $\mathbf{P}_{c,n-1}$ to its $b$th direction boundary position $\mathbf{P}_{b,n-1}$ on the equipotential line (cf. Fig.~\ref{fig:water_b}). $\theta_{{b},n-1}$ is the angle between lines $\overrightarrow{\mathbf{P}_{c,n-1}\mathbf{P}_{b,n-1}}$ and $\overrightarrow{\mathbf{P}_{c,n-1}\mathbf{P}^\prime_{c,n}}$, where $\mathbf{P}^\prime_{c,n}$ is the projected position of $\mathbf{P}_{c,n}$ on $\mathbf{P}_{c,n-1}$'s spatial plane, as in Fig.~\ref{fig:water_b}.

Basically, $r_{{b},n-1}^s$ in (\ref{equation:eqvelocity}) reflects the size variation of a water droplet, while $\theta_{b,n-1}$ evaluates the relative location of a boundary position $\mathbf{P}_{b,n-1}$ with respect to the motion direction $\overrightarrow{\mathbf{P}_{c,n-1}\mathbf{P}^\prime_{c,n}}$ of a water droplet. Since the size and motion direction of a water droplet is controlled by the shape and route of the 3D tube it passes through, by introducing $r_{{b},n-1}^s$ and $\theta_{b,n-1}$, a 3D tube's shape and route information can be effectively embedded.

Moreover, the terms $\frac{\lambda_1}{r_{b,n-1}^s}$ and $\cos\theta_{b,n-1}$ in (\ref{equation:eqvelocity}) reflect the impact of a 3D tube's shape to the viscosity and friction forces applied on a water droplet point $\mathbf{w}_b$. When $\mathbf{w}_{b}$ passes through a boundary position $\mathbf{P}_{b,n-1}$ located in front of the motion route of $\mathbf{w}_{b}$'s water droplet (i.e., small $\theta_{b,n-1}$), the friction force on $\mathbf{w}_{b}$ becomes large due to the increased normal force that $\mathbf{w}_{b}$'s water droplet applies on tube boundary $\mathbf{P}_{b,n-1}$. This will reduce the velocity of $\mathbf{w}_{b}$ and lead to larger $d_{b}^t$. Similarly, when a 3D tube becomes slim on $b$th direction (i.e., small $r_{b,n-1}^s$), the viscosity force on $\mathbf{w}_{b}$ will be enlarged which pulls $\mathbf{w}_{b}$ closer to $\mathbf{w}_{c}$ and creates smaller $d_{b}^t$. This point will be further discussed in the experimental results.

The third row of Fig.~\ref{fig:3D_tube} shows the droplet results derived from the 3D tubes in the second row of Fig.~\ref{fig:3D_tube} by using (\ref{equation:eq8}) and (\ref{equation:eqvelocity}). From Fig.~\ref{fig:3D_tube}, we can observe the effectiveness of our water-droplet process:

\squishlist
 \item The major motion directions of 3D tubes are properly captured by the large sectors in droplet results. For example, the water droplet of trajectory $C$ has a large sector in the upward direction since trajectory $C$ moves forward only. Comparatively, the water droplet of trajectory $B$ has large sectors in both top and left directions. This indicates the `forward+left turn' movement of $B$.
 \item Droplets derived from thick 3D tubes have larger sizes than those derived from slim tubes. For example, the droplet for trajectory $C$ has large size since $C$ follows a frequent motion pattern in the scene and has a thick 3D tube. Comparatively, since trajectory $A$ turns to an irregular region in the middle, its 3D tube becomes narrow in the later part. This leads to a small size in its corresponding droplet. Fig.~\ref{fig:3D_tube} implies that the size of a droplet can effectively differentiate regular and irregular motion patterns. Therefore, in this paper, droplet size is utilized as a major feature to detect abnormalities in trajectory analysis (cf.~(\ref{equation:eqab})).
\squishend

Finally, the obtained water droplet is sampled to achieve a low-dimensional droplet vector. In this paper, we simply concatenate time-axis distances $d_{b}^t$ in a water droplet as the low-dimensional droplet vector:
\begin{equation}
\mathcal{\mathbf{V}}^m=[d^{t,m}_{1},d^{t,m}_{2},\dots,d^{t,m}_{N_w}]
\label{equation:vector}
\end{equation}
where $\mathcal{\mathbf{V}}^m$ is the droplet vector for trajectory $\mathbf{\Gamma}^m$. $d^{t,m}_{b}$ is the time-axis distance in $b$th direction of $\mathbf{\Gamma}^m$'s water droplet. $N_w$ is the length of the vector and is set as $36$ in our experiments.

\section{Implementation in Trajectory Analysis \label{section:single_traj}}

With the tube-and-droplet representation, trajectories can be depicted by droplet vectors
and analyzed accordingly. In this section, we discuss the implementations of our tube-and-droplet representation in three trajectory analysis applications: trajectory clustering, trajectory classification \& abnormality detection, and 3D action recognition.

\subsection{Trajectory Clustering}

When performing trajectory clustering, we first utilize all trajectories being clustered as the given trajectories to construct thermal transfer fields (cf. Section~\ref{section:fields}). Then, a 3D tube and a droplet vector are derived based on the constructed thermal transfer fields to represent each trajectory (cf. Sections~\ref{section:3Dtube} and~\ref{section:droplet method}). Finally, we measure the distance between trajectories by calculating the distance between their corresponding droplet vectors, and perform trajectory clustering according to these droplet-vector distances. In this paper, we utilize Euclidean distance to measure the distance between droplet vectors, and utilize spectral clustering \cite{38} to cluster trajectories.

\subsection{Trajectory Classification \& Abnormality Detection \label{Section:Abnormal}}

In trajectory classification and abnormality detection, a set of normal training trajectories are provided whose class labels are given. We aim to recognize the classes of input test trajectories with the guidance of training trajectories, and identify abnormal test trajectories which do not follow the regular motion patterns provided by training trajectories.

Similar to trajectory clustering, we utilize all normal trajectories in the training set to construct thermal transfer fields, and derive a droplet vector for each individual training trajectory. Note that since our approach utilizes the \emph{complete} contextual information, we do not differentiate trajectory classes, that is, normal training trajectories from different trajectory classes are utilized indiscriminatively when constructing thermal transfer fields.

During testing, we first obtain a droplet vector for an input test trajectory. Then, the abnormality of a test trajectory is evaluated by its corresponding droplet vector. Since the size of a droplet can effectively differentiate regular and irregular
motion patterns, we detect a test trajectory $\mathbf{\Gamma}^m$ to be abnormal if:
\begin{equation}
\max_b{\{d^{t,m}_{b}\}}+\frac{1}{N_w}\sum_{b}{d^{t,m}_{b}}<TH
\label{equation:eqab}
\end{equation}
where $d^{t,m}_{b}$ is the $b$th element in $\mathbf{\Gamma}^m$'s droplet vector. ${N_w}$ is the length of the droplet vector (cf. (\ref{equation:vector})). $TH$ is a threshold decided by specific scenario. In the experiments of this paper, we simply calculate $\max_b{\{d^{t,m}_{b}\}}+\frac{1}{N_w}\sum_{b}{d^{t,m}_{b}}$ value for all normal trajectories in the training set, select the smallest value $T$ from them, and set 0.9$T$ as the threshold. In (\ref{equation:eqab}), the term $\frac{1}{N_w}\sum_{b}{d^{t,m}_{b}}$ is calculated to measure the size of a droplet while the term $\max_b{\{d^{t,m}_{b}\}}$ is used to evaluate the normality in a trajectory's major motion direction.

Finally, if a test trajectory $\mathbf{\Gamma}^m$ is evaluated as normal by (\ref{equation:eqab}), a one-against-all linear SVM classifier \cite{39} trained directly from the droplet vectors of training trajectories is applied to classify $\mathbf{\Gamma}^m$ into one of the trajectory classes.

\subsection{3D Action Recognition \label{section:3D action}}

\begin{figure}
  \centering
  \includegraphics[width=0.4\textwidth]{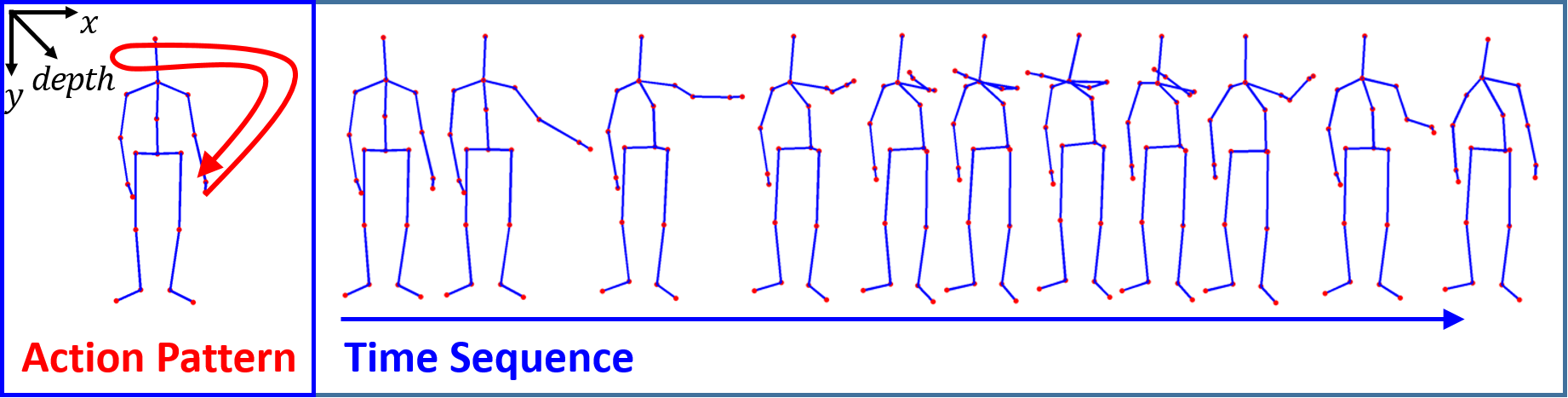}
  \caption{An example of 3D skeleton sequence from MSR-Action3D dataset. Left: The 3D trajectory of `horizontal wave' action of a `left hand' body point. Right: The skeleton sequence for `horizontal wave' action.}\label{fig:3D_skeleton_sequence}
\end{figure}

We also extend our tube-and-droplet approach into the application of 3D action recognition. In 3D action recognition, 3D skeleton sequences are provided which depict human actions in a 3D $x$-$y$-$depth$ space \cite{28,51,52,53,54}. An example skeleton sequence is shown in Fig.~\ref{fig:3D_skeleton_sequence}.

Since skeleton sequences are described by the trajectories of multiple body points (e.g., the red curve in Fig.~\ref{fig:3D_skeleton_sequence}), they are able to be represented and analyzed by the proposed tube-and-droplet approach. However, a 3D skeleton sequence differs from a regular motion trajectory in: 1) A 3D skeleton sequence includes multiple trajectories for different body points of a human; 2) The trajectory of a body point is located in 3D space ($x$-$y$-$depth$) instead of a 2D plane. Therefore, we extend our tube-and-droplet representation into higher dimension to depict 3D trajectories. Besides, in order to handle multiple trajectories in a 3D skeleton sequence, we represent each trajectory independently and then combine them together to achieve a unified depiction of the entire skeleton sequence.

The detailed implementation of our approach on 3D action recognition is described in the following.

First, thermal transfer fields are constructed for each body point based on the trajectories from training skeleton sequences. Since trajectories are located in 3D space, thermal transfer fields need to be modeled by 3D volumes. Therefore, we introduce two additional directions (i.e., $z^-$ and $z^+$ representing forward and backward directions in depth axis) and extend the process of constructing thermal transfer fields (cf. (\ref{equation:eqfield})) into 3D space. Furthermore, in order to make thermal transfer fields more powerful for different 3D actions, we construct a set of 3D thermal transfer fields for each individual action class under each body point. Therefore, if there are $N_\mathcal{A}$ action classes being recognized and $N_\mathcal{B}$ body points in a skeleton, a total of $N_\mathcal{A}\times N_\mathcal{B}$ sets of thermal transfer fields are constructed.

\begin{figure}
  \centering
  \vspace{-2mm}
  \subfloat[]{\includegraphics[height = 1.7 cm]{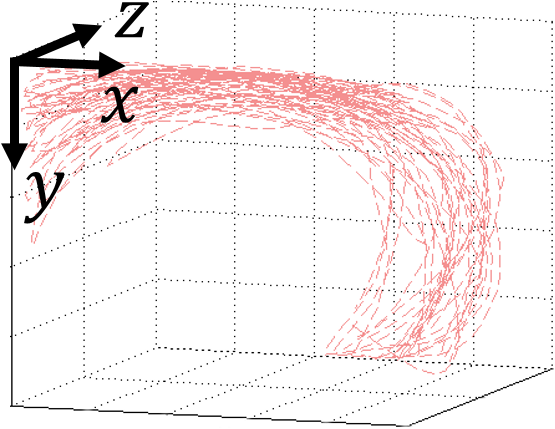}\label{fig:3D_traj}}
  \hspace{1mm}
  \subfloat[]{\includegraphics[height = 1.7 cm]{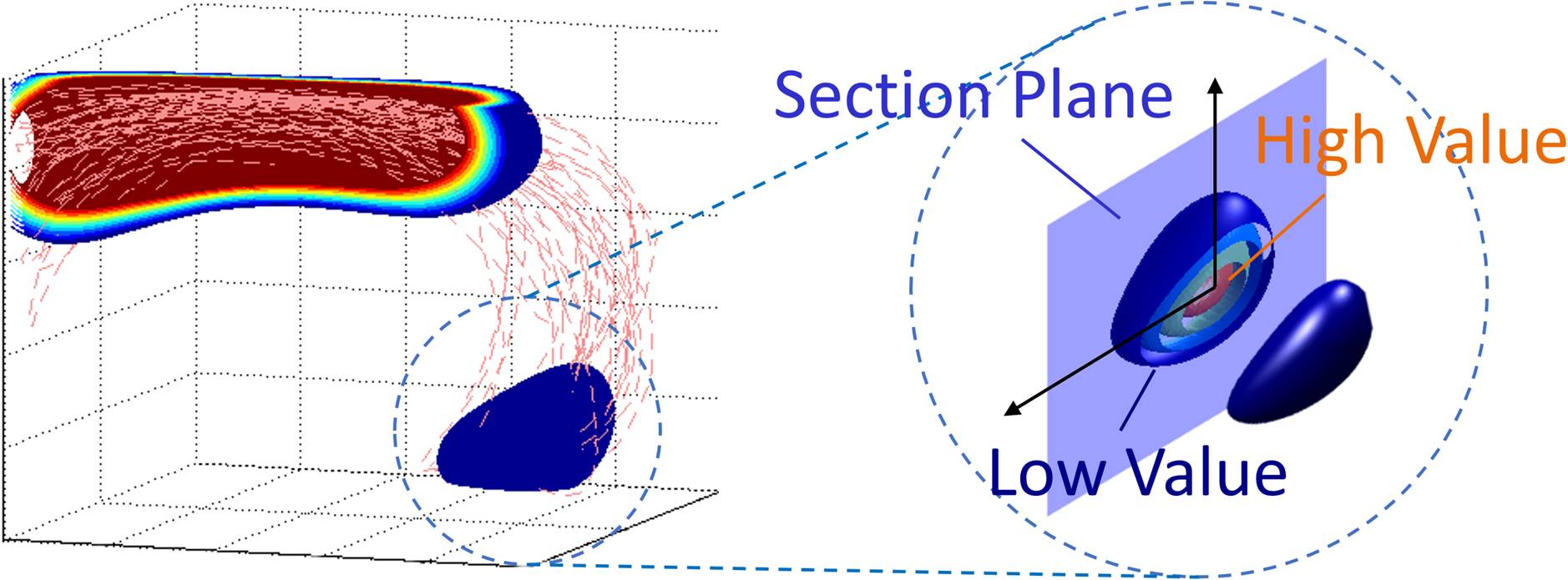}\label{fig:3D_field_detail1}}
  \\
  \subfloat[]{\includegraphics[height = 1.1 cm]{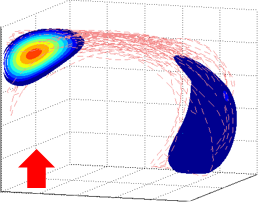}\label{fig:3D_field_1}}
  \hspace{0mm}
  \subfloat[]{\includegraphics[height = 1.1 cm]{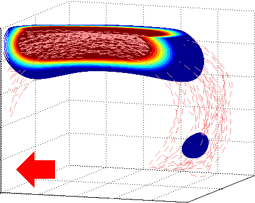}\label{fig:3D_field_2}}
  \hspace{0mm}
  \subfloat[]{\includegraphics[height = 1.1 cm]{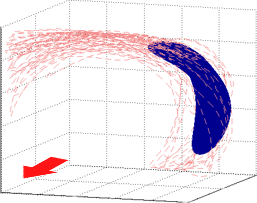}\label{fig:3D_field_3}}
  \hspace{0mm}
  \subfloat[]{\includegraphics[height = 1.1 cm]{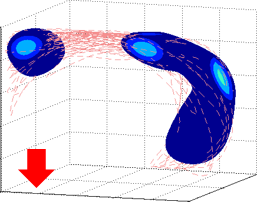}\label{fig:3D_field_4}}
  \hspace{0mm}
  \subfloat[]{\includegraphics[height = 1.1 cm]{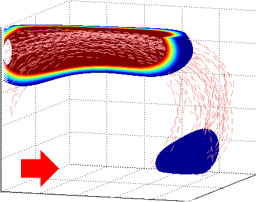}\label{fig:3D_field_5}}
  \hspace{0mm}
  \subfloat[]{\includegraphics[height = 1.1 cm]{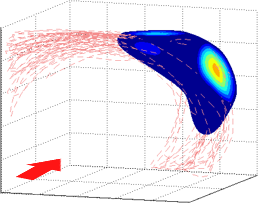}\label{fig:3D_field_6}}
  \caption{An example of thermal transfer fields for 3D trajectories. (a) Given 3D trajectories. (c)-(h) Constructed thermal transfer fields for the given 3D trajectories in (a). Note that (c)-(h) correspond to the thermal transfer fields in directions $y^-$, $x^-$, $z^-$, $y^+$, $x^+$, $z^+$, respectively. (b) An illustration of details inside a 3D thermal transfer field when cutting a high-value volume in a transfer field.}\label{fig:3D_field}
\end{figure}

Fig.~\ref{fig:3D_field} shows an example set of thermal transfer fields which represents the motion pattern of `horizontal wave' action for a `left hand' body point (cf. Fig.~\ref{fig:3D_skeleton_sequence}). According to Fig.~\ref{fig:3D_field}, six thermal transfer fields are constructed to describe global motion patterns in six directions inside a 3D space (cf. Figs.~\ref{fig:3D_field_1}-\ref{fig:3D_field_6}). Besides, each thermal transfer field is modeled as a 3D volume indicating the thermal propagation strength at different positions.

\begin{figure}
  \centering
  \includegraphics[width=0.4\textwidth]{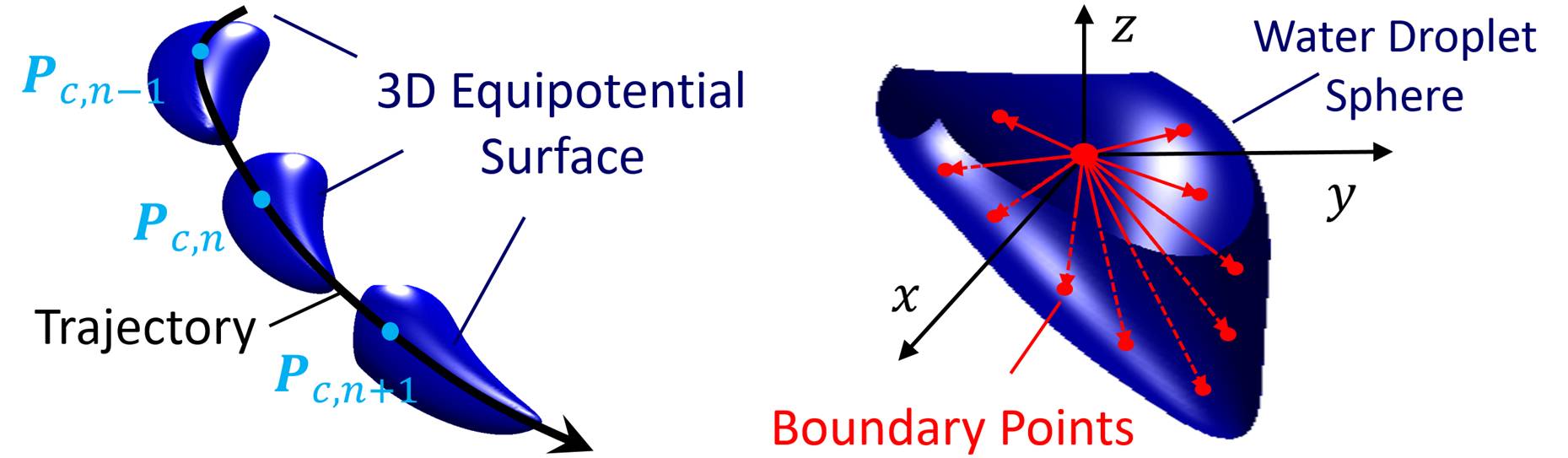}
  \caption{An example of equipotential surfaces (left) and the resulting water droplet sphere (right) for 3D trajectories.}\label{fig:3D_Equipotential_Plane}
\end{figure}

After thermal transfer fields are constructed, a tube and a droplet vector is constructed to depict the information of an input trajectory. Due to the 3D feature of input trajectories, the constructed tube is extended to a combination of 3D equipotential surfaces instead of 2D equipotential lines, as in the left figure of Fig.~\ref{fig:3D_Equipotential_Plane}. Similarly, the obtained droplet is also extended from a surface to a sphere which is represented by a center point and a set of boundary points located on a closed surface surrounding it (cf. right figure in Fig.~\ref{fig:3D_Equipotential_Plane}).

Finally, a skeleton sequence can be described by concatenating droplet vectors from different body points and for different action classes, as:
\begin{align}
\label{equation:3D_droplet}
\mathbf{{U}}^m=[&\omega_1\mathbf{V}^m_{1,1},\dots,\omega_{N_\mathcal{B}}\mathbf{V}^m_{N_\mathcal{B},1};\omega_1\mathbf{V}^m_{1,2},\dots,\omega_{N_\mathcal{B}}\mathbf{V}^m_{N_\mathcal{B},2};\dots \nonumber \\
                &\omega_1\mathbf{V}^m_{1,N_\mathcal{A}},\dots,\omega_{N_\mathcal{B}}\mathbf{V}^m_{N_\mathcal{B},N_\mathcal{A}}]
\end{align}
where $\mathbf{{U}}^m$ is the droplet-based feature vector for the $m$th skeleton sequence. $\mathbf{V}^m_{\beta,\alpha}$ is the droplet vector derived for $\beta$th body point and $\alpha$th action class. $\omega_\beta$ is the weighting factor balancing the relative importance of $\beta$th body point and it can be decided by cross-validation\cite{40}. $N_\mathcal{B}$ and $N_\mathcal{A}$ are the total number of body points and action classes.

During 3D action recognition, we first derive a droplet feature vector $\mathbf{{U}}^m$ for an input skeleton sequence, and then utilize a classifier to recognize the action class of this skeleton sequence. In this paper, we use two different classifiers 
: 1) KNN, and 2) one-against-all linear SVM classifier \cite{39}.

\section{Experimental Results\label{section:experiments}}

We evaluate the performance of our trajectory representation approach on three trajectory analysis applications: trajectory clustering, trajectory classification \& abnormality detection, and 3D action recognition. The experiments are performed on multiple benchmark trajectory datasets including Vehicle Motion Trajectory dataset (VMT)~\cite{10}, Simulated Traffic Intersection dataset (CROSS)~\cite{17}, our own constructed crossroad traffic dataset (TRAFFIC){\footnote {www.dropbox.com/s/ahyxw6vqypgb0uf/TRAFFIC.zip?dl=0}}, and MSR-Action3D Dataset (MSR) \cite{32}. $\lambda_1$ and $\lambda_2$ in (\ref{equation:eqvelocity}) are set as $2$ and $0.1$, which are decided from experimental statistics.

\subsection{Trajectory Clustering}

We perform trajectory clustering experiments on a benchmark Vehicle Motion Trajectory dataset (VMT)~\cite{10}. The VMT dataset includes $1500$ real-scene vehicle trajectories labeled in $15$ clusters. Some example trajectories in VMT dataset is shown in Fig.~\ref{fig:Hu_compare2}. In this experiment, we cluster trajectories into the same number of clusters as the ground truth (i.e., $15$ clusters), and evaluate the consistency between trajectory labels in the clustering results and those in the ground truth.

We compare our approach with seven methods: 1) Using Euclidean distance to measure the distance between trajectories plus using K-means clustering to cluster trajectories (\textbf{ED+Kmeans}); 2) Using Euclidean distance plus spectral clustering \cite{38} (\textbf{ED+SC}); 3) Using dynamic time warping to measure trajectory distances plus K-means \cite{41} (\textbf{DTW+Kmeans}); 4) Using DTW plus spectral clustering \cite{42} (\textbf{DTW+SC}); 5) Using a time-sensitive Dirichlet process mixture model \cite{10} to represent and cluster trajectories (\textbf{tDPMM}); 6) Using a 3-stage hierarchical learning model \cite{17} to represent and cluster trajectories (\textbf{3SHL}); 7) Using a heat-map model \cite{14} to represent and cluster (\textbf{HM}).

Note that the `tDPMM' and `3SHL' methods are state-of-the-art trajectory-modeling methods which increase the informativeness of trajectories by constructing probability models for each trajectory class. Besides, the `HM' method encodes the temporal variation of a trajectory.

Moreover, in order to evaluate the effectiveness of our water droplet process, we further include the results of three additional methods: 1) Using our approach to construct 3D tubes for trajectory representation, plus using Hausdorff distance \cite{23} to capture the high-dimensional information in these 3D tubes for trajectory clustering (\textbf{3D Tube+Hausdorff}); 2) Using 3D tubes plus using a state-of-the-art Grassmann manifold method \cite{24} (\textbf{3D Tube+Manifold}); 3) Using our approach to achieve thermal diffusion maps for each trajectory point (cf. Fig.~\ref{fig:rabc}), and directly concatenating these thermal diffusion maps as the representation of a trajectory, finally using the Grassmann manifold method~\cite{24} to capture the high-dimensional information (\textbf{Thermal map+Manifold}).

Note that the major difference between the `Thermal map+Manifold' method and `3D Tube+Manifold' method is that `Thermal map+Manifold' skips the step of equipotential line extraction (cf. Section~\ref{section:equilines}), and directly utilizes a thermal diffusion map to represent a trajectory point.

\subsubsection{Comparison of clustering results}

Table~\ref{table:table_Hu_ori} compares the cluster learning accuracy~\cite{10} for different methods on VMT dataset, where the cluster learning accuracy measures the total percentage of trajectories being correctly clustered. From Table~\ref{table:table_Hu_ori}, approaches using our 3D tube representation (3D Tube+Hausdorff, 3D Tube+Manifold, Thermal map+Manifold, Ours) achieve obviously better clustering results than the compared methods. This demonstrates the usefulness of our 3D tube representation. Moreover, we can also observe from Table~\ref{table:table_Hu_ori} that: 1) Our approach, which integrates both 3D tube representation and water droplet process, achieves the best clustering results. This demonstrates the effectiveness of our tube+droplet framework. 2) The `3D Tube+Manifold' method has slightly better results than the `Thermal map+Manifold' method. This implies that equipotential lines (cf. Section~\ref{section:equilines}) can not only capture the useful information in thermal diffusion maps, but also suitably avoid the disturbance from noise in thermal diffusion maps.

\begin{table}
  \centering
  \caption{Cluster Learning Accuracy for different methods on VMT Dataset (\%)}\label{table:table_Hu_ori}
  \begin{tabular}{lrr}
    \hline
    Method                              &Cluster Accuracy              \\ \hline
    ED+Kmeans                               &82.6                     \\
    ED+SC\cite{38}                                   &85.0                      \\
    DTW+Kmeans\cite{41}                             &83.2                       \\
    DTW+SC\cite{42}                                  &85.3                     \\
    tDPMM~\cite{10}                         &86.7                       \\
    3SHL~\cite{17}                             &84.4                       \\
    HM~\cite{14}                              &82.0                       \\ \hline
    \textbf{3D Tube+Hausdorff}              &91.5                       \\
    \textbf{Thermal Map+Manifold}           &92.2         \\
    \textbf{3D Tube+Manifold}              &93.6                         \\
    \textbf{Ours}                           &\textbf{93.8}              \\
    \hline
  \end{tabular}
\end{table}

\subsubsection{Robustness to noises and trajectory breaks}

We further demonstrate the effectiveness of our approach in dealing with noisy or broken trajectories. Following~\cite{10}, we add Gaussian noise to all points in a trajectory to simulate a noisy trajectory. Three noise levels are used to derive three trajectory datasets with different noise strengths (i.e., Noise Level $1$, $2$, and $3$ in Table~\ref{table:table_Hu_noise&omit}). Similarly, we omit the initial or last $G$ points in two of $10$ trajectories in each cluster to simulate datasets with broken trajectories (i.e., omit $G=10\%$, $G=20\%$, $G=30\%$, and $G=40\%$ in Table~\ref{table:table_Hu_noise&omit})~\cite{10}.

\begin{table*}[t]
  \newcommand{\tabincell}[2]{\begin{tabular}{@{}#1@{}}#2\end{tabular}}
  \centering
  \caption{Cluster Learning Accuracy with different noise or trajectory break levels on VMT Dataset (\%)}
  \begin{tabular}{
    @{\hspace{3pt}}c@{\hspace{3pt}} |
    @{\hspace{3pt}}c@{\hspace{3pt}} |
    @{\hspace{3pt}}c@{\hspace{3pt}} |
    @{\hspace{3pt}}c@{\hspace{3pt}} |
    @{\hspace{3pt}}c@{\hspace{3pt}} |
    @{\hspace{3pt}}c@{\hspace{3pt}} |
    @{\hspace{3pt}}c@{\hspace{3pt}} |
    @{\hspace{3pt}}c@{\hspace{3pt}} |
    @{\hspace{3pt}}c@{\hspace{3pt}} |
    @{\hspace{3pt}}c@{\hspace{3pt}} |
    @{\hspace{3pt}}c@{\hspace{3pt}}
    }
    \hline
    Datasets &\tabincell{c}{ED+Kmeans} &\tabincell{c}{ED+SC \\ ~\cite{38}}
    &\tabincell{c}{DTW+Kmeans \\ ~\cite{41}} &\tabincell{c}{DTW+SC \\ ~\cite{42}}
    &\tabincell{c}{tDPMM \\ ~\cite{10}} &\tabincell{c}{3SHL \\ ~\cite{17}} &\tabincell{c}{HM \\ ~\cite{14}} &\tabincell{c}{3D Tube+ \\ Hausdorff} &\tabincell{c}{3D Tube+ \\ Manifold} &\textbf{Ours} \\ \hline
    Noise Level 1 &80.6      &83.3  &81.2       &83.5   &84.3    &83.2   &80.3   &88.7 &91.3   & \textbf{91.7}\\
    Noise Level 2 &78.8      &81.2  &80.0       &81.9   &83.3    &80.1   &78.9   &85.3 &89.1   & \textbf{90.7}\\
    Noise Level 3 &77.0      &81.2  &78.8       &74.6   &81.5    &79.2   &71.8   &79.5 &86.4   & \textbf{88.1}\\ \hline
    Omit $G$=10\% &80.8      &84.4  &81.1       &85.1   &86.1    &84.0   &82.0   &90.6 &92.9   & \textbf{93.1}\\
    Omit $G$=20\% &78.9      &82.7  &76.5       &83.5   &85.7    &80.4   &81.8   &89.7 &91.3   & \textbf{92.1}\\
    Omit $G$=30\% &76.5      &79.3  &74.4       &81.1   &81.8    &76.7   &80.3   &87.2 &88.7   & \textbf{89.5}\\
    Omit $G$=40\% &72.3      &76.1  &71.8       &77.5   &78.1    &74.1   &78.1   &84.3 &83.2   & \textbf{87.2}\\
    \hline
  \end{tabular}\label{table:table_Hu_noise&omit}
\end{table*}

Table~\ref{table:table_Hu_noise&omit} compares the clustering results of different methods on noisy or broken trajectories derived from VMT dataset. Table~\ref{table:table_Hu_noise&omit} shows that our approach achieves the best clustering results under different noise or trajectory break levels. Moreover, when noise or trajectory break level increases, the clustering performance decrease by our approach is relatively small among the compared methods, this further demonstrates the robustness of our tube-and-droplet approach when handling noises or trajectory breaks.

\subsubsection{Effectiveness of 3D tube representation}

\begin{figure}
  \centering
  \subfloat[]{\includegraphics[height = 2.8 cm]{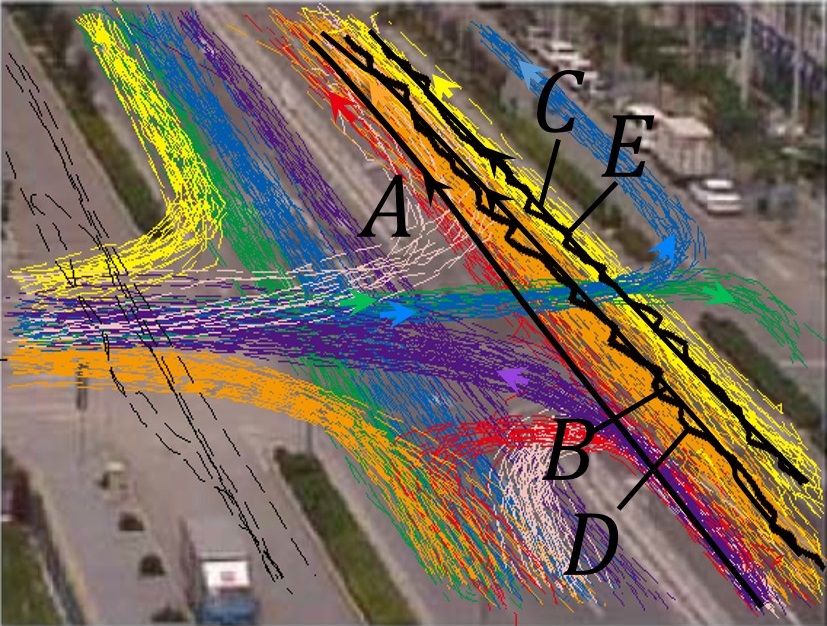}\label{fig:Hu_traj2}}
  \hspace{1mm}
  \subfloat[]{\includegraphics[height = 2.8 cm]{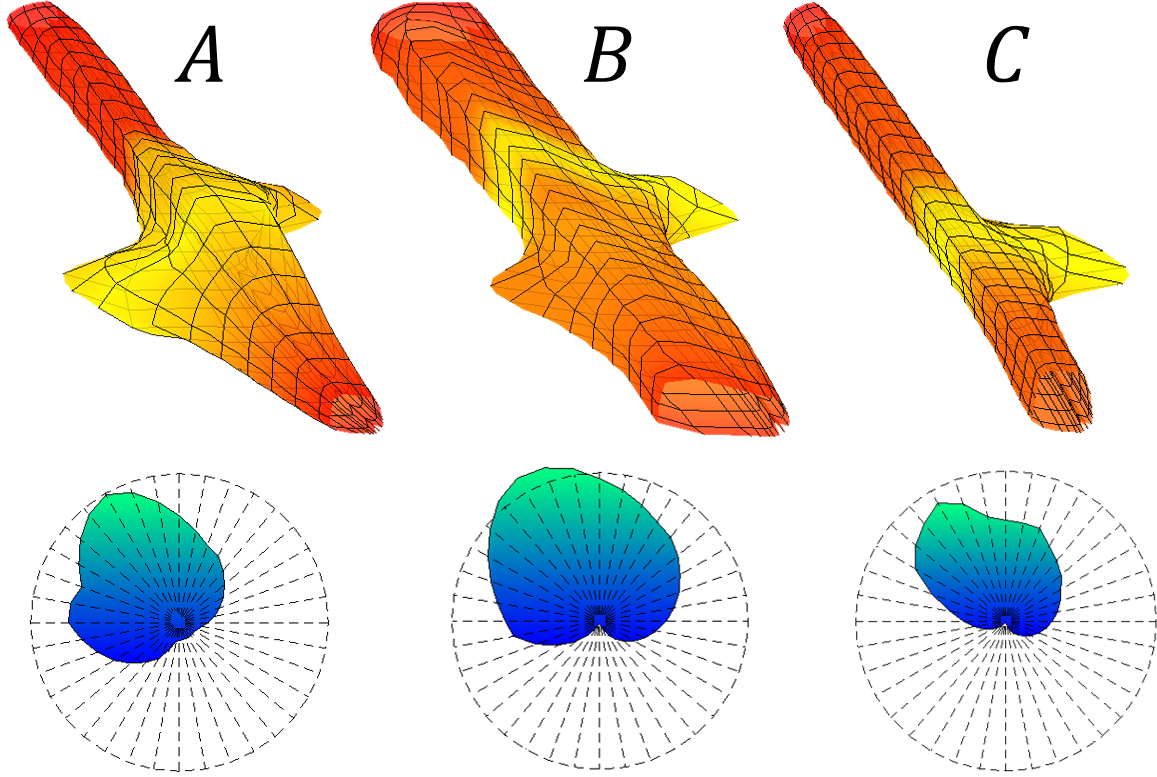}\label{fig:Hu_compare_1}}\\
  \subfloat[]{\includegraphics[height = 3 cm]{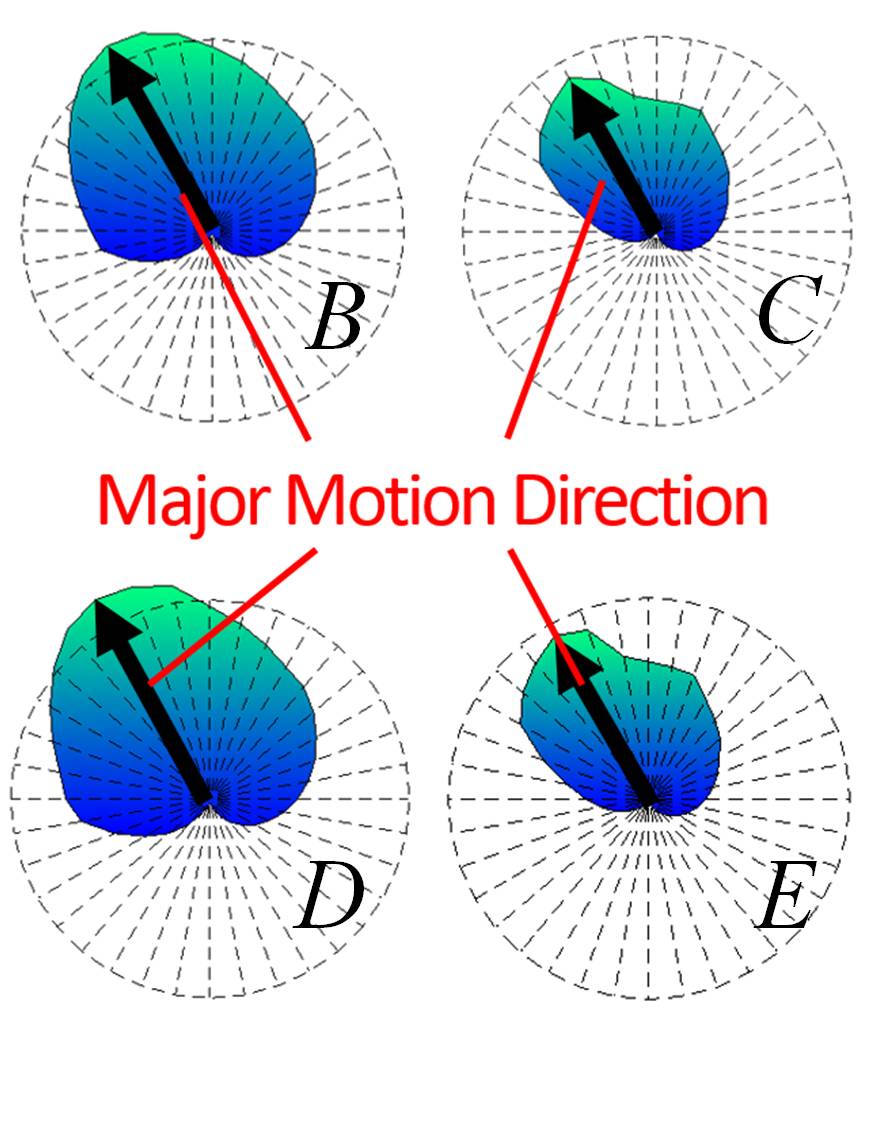}\label{fig:Hu_noise_4}}
  \hspace{3mm}
  \subfloat[]{\includegraphics[height= 3 cm]{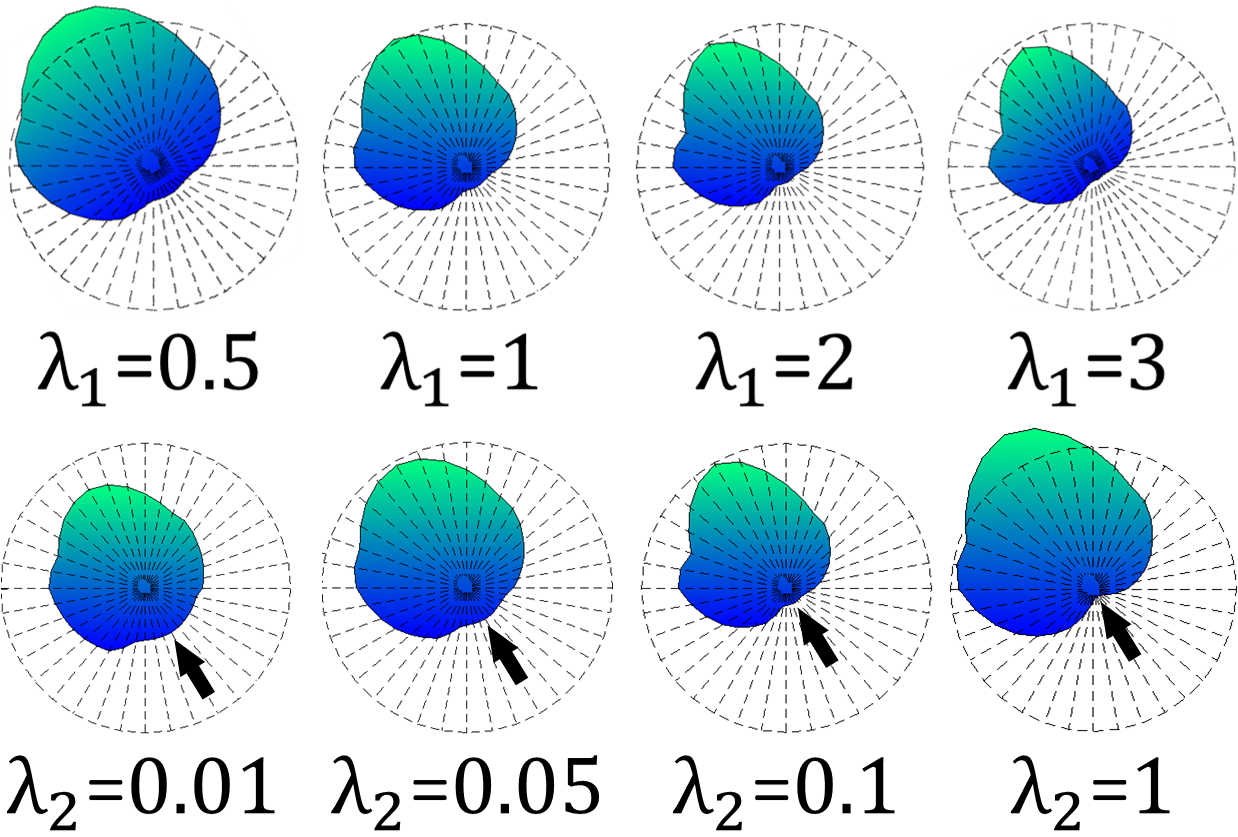}\label{fig:drop_final_a}}
  \caption{(a) Trajectories from different groundtruth clusters in VMT dataset (color curves) and five input trajectories (black curves labeled by $A$-$E$). (b) Upper: 3D tubes for trajectories $A$, $B$, and $C$ in (a); Lower: Water droplets for $A$, $B$, and $C$. (c) Comparison of water droplet results between clean trajectories ($B$, $C$) and noisy trajectories ($D$, $E$) in (a) (note that $B$, $D$ are from one cluster and $C$, $E$ are from another cluster). (d) Upper: Droplets for trajectory $A$'s 3D tube in Fig.~\ref{fig:Hu_compare_1} under different $\lambda_1$ where $\lambda_2=0.1$; Lower: Droplets for trajectory $A$'s 3D tube in Fig.~\ref{fig:Hu_compare_1} under different $\lambda_2$ where $\lambda_1=2$. (Best viewed in color)}\label{fig:Hu_compare2}
\end{figure}

Figs~\ref{fig:Hu_compare2} illustrates a more detailed example about the effectiveness of our 3D tube representation. In Fig.~\ref{fig:Hu_traj2}, the colored curves are trajectories to be clustered and the black trajectories labeled in $A$, $B$, and $C$ are three trajectories from them. In order to ease the discussion, trajectories from different groundtruth clusters are displayed by different colors.
Since trajectories $A$, $B$, and $C$ come from three trajectory clusters which are located close to each other and have similar motion patterns (cf. the clusters in yellow, orange, and red in Fig.~\ref{fig:Hu_traj2}), existing methods have limitations in differentiating them which only consider the pair-wise correlation between trajectories (ED+Kmeans, ED+SC, DTW+Kmeans, DTW+SC, HM) or intra-cluster correlation among trajectories (tDPMM, 3SHL).

With our 3D tube representation, the differences among $A$, $B$, and $C$ can be properly highlighted by embedding the complete motion information from all trajectories, as the upper figures in Fig.~\ref{fig:Hu_compare_1}. For example, the 3D tube of trajectory $A$ shows an obvious leftward convex part since there is a large left-turn pattern provided by the purple cluster along $A$'s route. Trajectory $B$'s tube is thicker since it is located in the middle of a large upleft-ward motion pattern jointly provided by the yellow, orange, and red clusters. Besides, trajectory $C$'s tube includes an obvious rightward convex part due to the rightward contextual patterns provided by the blue and green clusters next to $C$. Therefore, by suitably capturing the high-dimensional information in these 3D tubes, the difference among trajectories can be effectively reflected in the resulting droplet vectors (cf. the lower figures in Fig.~\ref{fig:Hu_compare_1}).


\subsubsection{Effectiveness of water droplet process}

In order to evaluate our water droplet process, we compare our approach with `3D Tube+Hausdorff', `3D Tube+Manifold', and `Thermal Map+Manifold'. These methods use the same 3D tube representation to depict a trajectory, but use different schemes to capture the high-dimensional information in a 3D tube. The clustering results of these methods under different noise or trajectory break levels are shown in Table~\ref{table:table_Hu_noise&omit}. Furthermore, Table~\ref{table:run_time} compares the running time of the 3D tube information handling \& clustering steps in these methods. We observe that:
\squishlist
 \item From Table~\ref{table:table_Hu_noise&omit}, the clustering-accuracy difference between our approach and `3D Tube+Hausdorff' becomes larger for higher noise levels. This implies that satisfactory results cannot be easily achieved without suitably capturing the high-dimensional information in 3D tubes. More specifically, since Hausdorff distance is easily affected by noise, its results decrease more rapidly for large noises. In contrast, since our water droplet process characterizes a 3D tube by accumulating information at different parts in a tube (cf. (\ref{equation:eq8})), the disturbance from noise can be effectively reduced.
 \item Although `3D Tube+Manifold' and `Thermal Map+ Manifold' can achieve relatively better results than `3D Tube+Hausdorff', their computation complexities are considerably high (cf. Table~\ref{table:run_time}). Comparatively, our water droplet process handles information of a 3D tube in a simple but effective way, which is able to perform clustering in less than 1 minute while achieving better results than `3D Tube+Manifold' and `Thermal Map+Manifold'.
\squishend

\begin{table}
 \centering
  \caption{Running time of different steps for Table~\ref{table:table_Hu_noise&omit}  \protect\\(Note: `Transfer field+3D tube' refers to the steps of constructing thermal transfer fields and 3D tubes. `Handling 3D tube+clustering' refers to the steps of handling the high-dimensional information in 3D tubes and performing clustering. Implemented by Matlab and evaluated on a PC with 4 core CPU and 8G RAM.)}
  \begin{tabular}{{c | c | c}}
    \hline
    Method                  &Transfer field             &Handling 3D tube      \\
                            &+3D tube                   &+clustering            \\  \hline
    3D Tube+Hausdorff       & \multirow{4}{*}{11.2 min}    & 38.4 min                    \\
    3D Tube+Manifold        &                           & 12.0 hr                    \\
    Thermal Map+Manifold    &                           & 12.0 hr                      \\
    \textbf{Ours}           &                           &\textbf{47.5 sec}   \\
    \hline
  \end{tabular}\label{table:run_time}
\end{table}

Moreover, Fig. \ref{fig:Hu_noise_4} shows the droplet results for four trajectories where trajectories $B$ and $D$ belong to the orange cluster and trajectories $C$ and $E$ belong to the yellow cluster. Since trajectories $D$ and $E$ are interfered by noise, the ambiguity among trajectories increases. However, with our water droplet process, these noisy effects can be properly reduced and the major characteristics of a 3D tube can be properly obtained. For example, in Fig.~\ref{fig:Hu_noise_4}, the common characteristics of trajectories $B$ and $D$ are captured in their droplets which include larger sectors on both sides of the trajectories' major motion direction.


\subsubsection{Effect of different parameter values}

Fig.~\ref{fig:drop_final_a} shows the droplet results for trajectory $A$'s 3D tube in Fig.~\ref{fig:Hu_compare_1} under different $\lambda_1$ and $\lambda_2$ values (cf. (\ref{equation:eqvelocity})).

From Fig.~\ref{fig:drop_final_a}, we can see that $\lambda_1$ mainly controls the effect from a 3D tube's shape. More specifically, when $\lambda_1$ increases, a droplet becomes more concentrated on a tube's true motion directions and convex parts, while sectors not along a tube's direction or convex part will shrink. At the same time, with the increase of $\lambda_1$, the thickness of a tube is also reflected more obviously in a droplet. For example, since trajectory $A$'s 3D tube in Fig.~\ref{fig:Hu_compare_1} is relatively narrow, increasing $\lambda_1$ will decrease the size of its droplet and make the droplet more coherent with the tube's thickness.

Besides, Fig.~\ref{fig:drop_final_a} reveals that $\lambda_2$ controls the impact of a 3D tube's route. When $\lambda_2$ is extremely small, the route information of a 3D tube cannot be fully included in a droplet (i.e., a droplet cannot tell whether a trajectory is moving upward or downward along a 3D tube, cf. Fig.~\ref{fig:Hu_compare_1}). This will create obvious sectors in the opposite direction to a trajectory's motion route (cf. the leftmost figure in Fig.~\ref{fig:drop_final_a}). When $\lambda_2$ increases, the route information is more clearly included in its corresponding droplet, and sectors opposite to a trajectory's route will shrink. We set $\lambda_1$ and $\lambda_2$ as $2$ and $0.1$. Experiments show that these values can properly embed both the shape and route information of a 3D tube and create satisfactory results.

\subsection{Trajectory Classification \& Abnormality Detection}

\subsubsection{Results on TRAFFIC dataset}

\begin{figure}
  \centering
  \subfloat[]{\includegraphics[height = 2.15 cm]{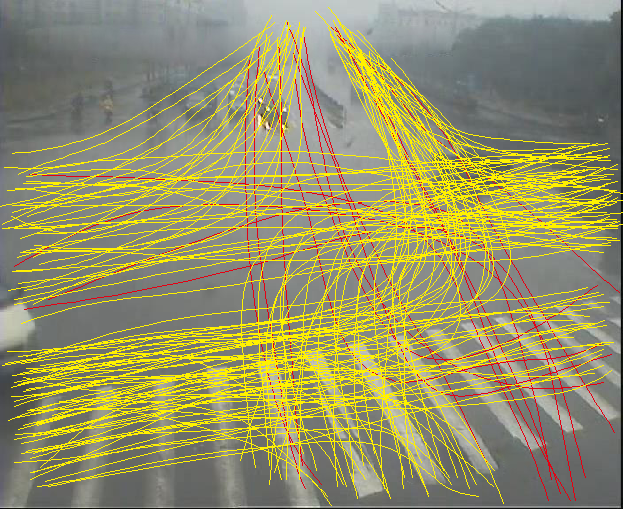}\label{fig:Our_abnormal_traj}}
  \hspace{0.5mm}
  \subfloat[]{\includegraphics[height = 2.15 cm]{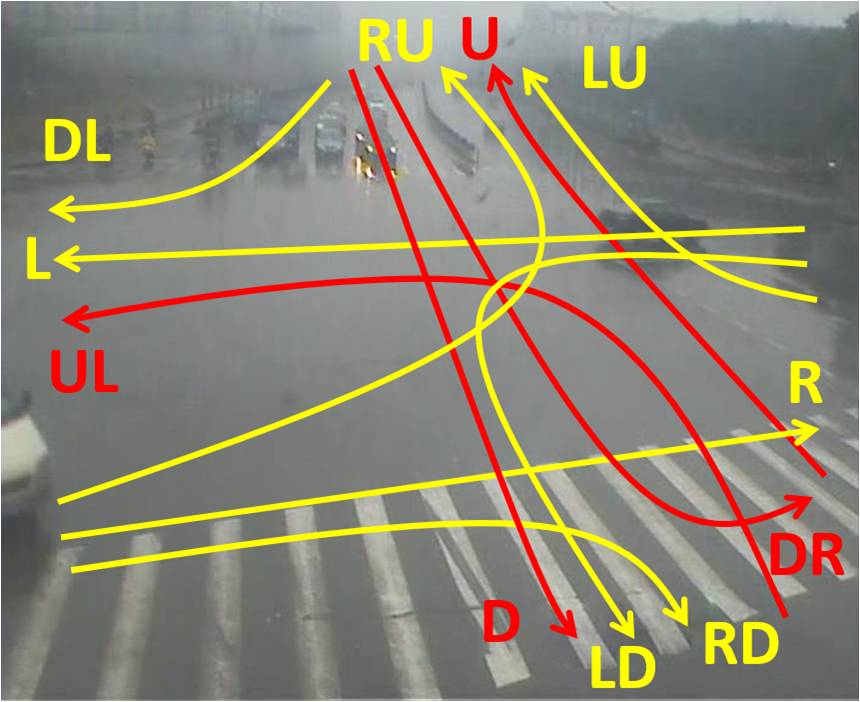}\label{fig:Our_abnormal_label}}
  \hspace{1mm}
  \subfloat[]{\includegraphics[height = 2.15 cm]{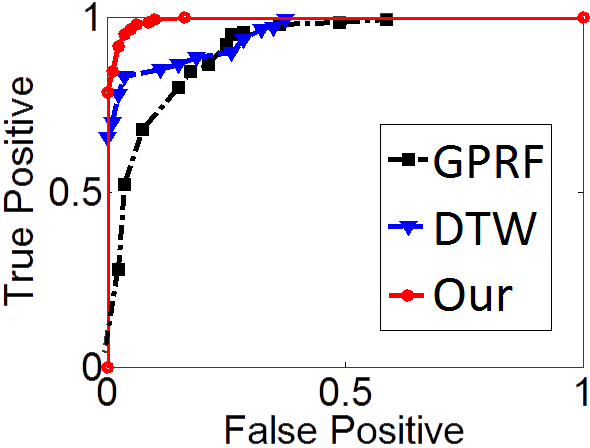}\label{fig:Our_abnormal_ROC}}
  \caption{(a) Example trajectories of the TRAFFIC dataset (yellow curves: normal trajectories, red curves: abnormal trajectories). (b) Major motion routes of different trajectory classes. The yellow and red curves indicate routes for normal and abnormal trajectory classes (`U',`D',`L',`R' refers to `upward',`downward',`leftward',`rightward' respectively. For example, `LU' means moving leftward and then upward). (c) ROC curves of different methods when discriminating normal/abnormal trajectories under different abnormality detection thresholds. (Best viewed in color)}\label{fig:Our_abnormal}
\end{figure}

We perform experiments of trajectory classification \& abnormality detection on our own constructed TRAFFIC dataset. This dataset includes $300$ real-scene trajectories where $200$ trajectories are for normal activities and the other $100$ trajectories are abnormal ones.
The normal trajectories includes seven classes, with about $30$ trajectories for each class.
The major motion routes of different trajectory classes are indicated in Fig.~\ref{fig:Our_abnormal_label}. Some example normal and abnormal trajectories are shown in Fig.~\ref{fig:Our_abnormal_traj}.

Note that this is a challenging dataset in that: 1) The total number of trajectories in the dataset is small, making it difficult to construct reliable models; 2) The motion trajectories within the same class have large variations due to the large width of roads; 3) Due to the visual angle of the surveillance camera, trajectories from different class are easily confused and are difficult to differentiate.

We compare our approach (cf. Sections~\ref{Section:Abnormal}) with two methods: 1) The GPRF method~\cite{11} which introduces Gaussian process regression flows to model the location and velocity probability for each trajectory class, and utilizes them to classify trajectories (\textbf{GPRF}); 2) The DTW method \cite{41} which classifies a test trajectory by measuring its dynamic-time-warping distance with the center of different trajectory classes (\textbf{DTW}).

\begin{figure}
  \centering
  \subfloat[]{\includegraphics[height = 2.35 cm, width=2.75 cm]{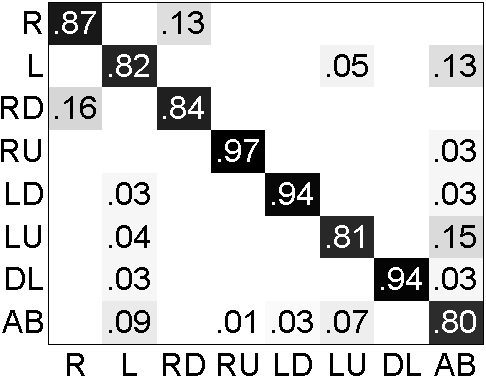}\label{fig:Our_abnormal_1}}
  \hspace{2mm}
  \subfloat[]{\includegraphics[height = 2.35 cm, width=2.75 cm]{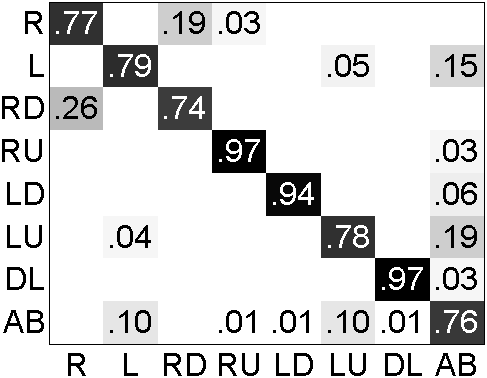}\label{fig:Our_abnormal_2}}
  \hspace{2mm}
  \subfloat[]{\includegraphics[height = 2.35 cm, width=2.75 cm]{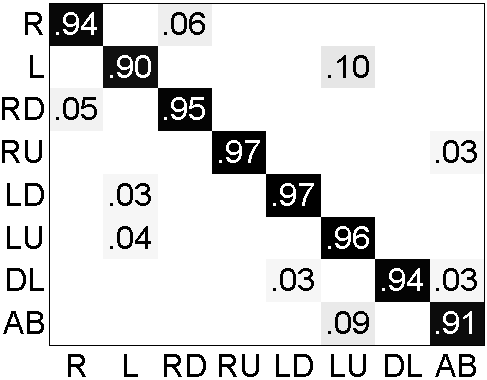}\label{fig:Our_abnormal_3}}
  \caption{Confusion matrices of different methods when classifying normal trajectories and detecting abnormal ones (denoted as `AB'). (a) GPRF~\cite{11}. (b) DTW~\cite{41}. (c) Ours.}\label{table:table_Our} 
\end{figure}

We split the dataset into $50\%$ training-$50\%$ testing parts. Note that in our experiments, only normal trajectories are used for training. Four independent runs are performed where the training and testing sets are randomly selected in each run, and the final results are averaged. Fig.~\ref{table:table_Our} compares the confusion matrices of different methods for classifying normal trajectories and detecting abnormal ones. Fig.~\ref{fig:Our_abnormal_ROC} further compares the ROC curves of different methods when discriminating normal/abnormal trajectories under different abnormality detection thresholds.

From Figs~\ref{table:table_Our} and~\ref{fig:Our_abnormal_ROC}, we can observe that:
\squishlist
\item Our approach can achieve obviously better results than the compared methods. More specifically, the compared methods have low effectiveness in discriminating trajectory classes such as \emph{RD}, \emph{L}, and \emph{LU}. This is because trajectories in these classes are easily confused with similar trajectories from other classes such as \emph{R} and \emph{AB} (e.g., \emph{L}, \emph{LU} are similar to patterns \emph{UL} and \emph{U} in the abnormal class \emph{AB}, cf. Fig.~\ref{fig:Our_abnormal_label}). Comparatively, since our approach introduces informative tube-and-droplet representation to capture the subtle difference between trajectory groups, more satisfactory results can be achieved.
\item Although the GPRF method constructs probability models to encode the spatial-temporal variation of trajectory classes, it still creates less satisfactory results. This is because: (a) The GPRF method only focuses on modeling the contextual information inside each trajectory class, which has limitations in differentiating trajectories from similar classes; (b) The number of training trajectories is small, which makes it difficult for the GPRF method to construct reliable probability models. Comparatively, our approach is able to work reliably under relatively small trajectories. Besides, by leveraging the complete contextual motion information, the subtle difference between different classes are also properly discriminated by our approach, thus obtaining more improved results.
\squishend

\subsubsection{Results on CROSS dataset}

We also perform experiments of trajectory classification and abnormality detection on CROSS dataset~\cite{17}. It includes $11400$ normal trajectories labeled in $19$ clusters and $200$ abnormal trajectories (one example in Fig.~\ref{fig:framework}).

Following~\cite{17}, we utilize $1900$ normal trajectories in training. The constructed thermal transfer fields or trajectory class models are then utilized to classify the rest $9500$ normal and $200$ abnormal trajectories. We use (\ref{equation:eqab}) to detect abnormal trajectories and a simple K-nearest neighbor (KNN)\cite{43} strategy to classify normal trajectories.

\begin{table}
  \centering
  \caption{Trajectory classification \& abnormality detection results on CROSS (\%)}
  \begin{tabular}{lcc}
    \hline
    Method                               &Classification     &Abnormality Detection  \\
                                          & (CA)                     &(DR/FPR)\\     \hline
    DTW~\cite{41}                                  &95.6                    &71.3/55.5  \\
    GPRF~\cite{11}                           &98.0                     &80.5/23.5  \\
    tDPMM~\cite{10}                         &98.0                     &91.0/\textbf{23.3}  \\
    3SHL~\cite{17}                             &96.8                    &85.5/23.5  \\
    \textbf{Ours}                           &\textbf{98.6}             &\textbf{91.3}/23.5  \\
    \hline
  \end{tabular}\label{table:table_CROSS_abnormal}
\end{table}

Table~\ref{table:table_CROSS_abnormal} compares the trajectory classification \& abnormality detection results. Classification results are evaluated by classification accuracy (CA) which is the ratio between total number of correctly classified normal trajectories and total number of normal trajectories. Abnormality detection results are evaluated by abnormality detection rate (DR) and abnormality false positive rate (FPR)~\cite{17}. Note that besides the DTW and GPRF methods, we also include the results of two state-of-the-art methods on CROSS dataset in Table~\ref{table:table_CROSS_abnormal} (i.e., 3SHL~\cite{17} and tDPMM~\cite{10}).

Our approach achieves the best performance in classification. When detecting abnormal trajectories, our approach can also achieve obviously improved results than the compared methods (DTW \cite{41}, 3SHL~\cite{17}, and GPRF~\cite{11}) and similar results to a state-of-the-art tDPMM method~\cite{10}.

\subsection{3D Action Recognition}

Finally, we evaluate the performance of our approach in 3D action recognition. We perform experiments on a benchmark MSR-Action3D Dataset \cite{32} which includes $557$ 3D skeleton sequences for $20$ human actions performed by $10$ different subjects. One example skeleton sequence is shown in Fig.~\ref{fig:3D_skeleton_sequence}.

Following the previous works on MSR-Action3D dataset \cite{32,55}, we evaluate the recognition accuracy over all $20$ actions where actions of half of the subjects are used for training and the rest actions are used for testing. Besides, a trajectory alignment process similar to~\cite{28} is applied as a pre-processing step to reduce 3D trajectory variations.

We utilize the process in Section~\ref{section:3D action} to implement our approach for 3D action recognition, where `Droplet+KNN' and `Droplet+SVM' in Table~\ref{table:table_MSR3D} refer to using KNN and SVM classifiers to recognize our droplet feature vectors (cf. (\ref{equation:3D_droplet})), respectively. Moreover, we also include the results by combining our droplet feature vector with a state-of-the-art `Moving Poselets' method~\cite{59} which introduces sophisticated mid-level classifiers to improve recognition accuracy (cf. `Droplet+Moving Poselets' in Table~\ref{table:table_MSR3D}). Specifically, we concatenate our droplet feature vectors with the body point velocity \& acceleration features used in~\cite{59}, and follow the `Moving Poselets' classification process~\cite{59} to recognize the action class of the concatenated feature vectors.

We compare our approach with the state-of-the-art 3D action recognition methods using skeleton sequences \cite{56,55,51,52,53,54,57,59}. Table~\ref{table:table_MSR3D} shows the recognition accuracy. According to Table~\ref{table:table_MSR3D}, our `Droplet+SVM' approach outperforms all the existing techniques except~\cite{59}. This demonstrates that our tube-and-droplet framework can be reliably applied to handle sequence analysis with multiple trajectories. Besides, our `Droplet+KNN' approach also achieves satisfactory results. It implies that our droplet features can effectively capture the discriminative characteristics of trajectories, such that good results can be achieved with simple recognition strategies such as KNN. Moreover, the `Droplet+Moving Poselets' approach achieves the best performance. It further indicates that our droplet features can be effectively combined with more sophisticated recognition strategies to achieve further improved performances.

\begin{table}
  \centering
  \caption{Recognition accuracy comparison on MSR-Action3D dataset (\%)}
  \begin{tabular}{lrr}
    \hline
    Method                              &Recognition Accuracy              \\ \hline
    Eigen joints~\cite{56}                             &82.3                       \\
    HON4D+$D_{disc}$~\cite{55}                             &88.9                       \\
    Actionlet Ensemble~\cite{51}               &88.2                      \\
    Skeleton Quads~\cite{53}                                  &89.9                     \\
    Pose Set~\cite{57}                                        &90.2            \\
    Manifold Learning~\cite{54}                                  &91.2                      \\
    Moving Pose~\cite{52}                                  &91.7   \\
    Moving Poselets~\cite{59}            &93.6
\\\hline
    \textbf{Droplet+KNN}                       &91.2                        \\
    \textbf{Droplet+SVM}                              &92.1         \\
    \textbf{Droplet+Moving Poselets}              &\textbf{93.9}\\
    \hline
  \end{tabular}\label{table:table_MSR3D}
\end{table}

\section{Conclusion\label{section:conclusion}}

In this paper, we study the problem of informative trajectory representation and introduce a novel tube-and-droplet framework. The framework consists of three key ingredients: 1) introducing the idea of constructing thermal transfer fields to embed the global motion patterns in a scene; 2) deriving equipotential lines and concatenating them into a 3D tube to establish a highly informative representation, which properly embeds both the motion route and the contextual motion pattern for a trajectory; 3) introducing a simple but effective droplet-based process to effectively capture the rich information in 3D tube representation. We apply our tube-and-droplet approach to various trajectory analysis applications including clustering, abnormality detection, and 3D action recognition. Extensive experiments on benchmark demonstrate the effectiveness of our approach.

\bibliographystyle{IEEEtran}
\bibliography{egbib}

\begin{IEEEbiography}[{\includegraphics[width=1in,height=1.2in,clip,keepaspectratio]{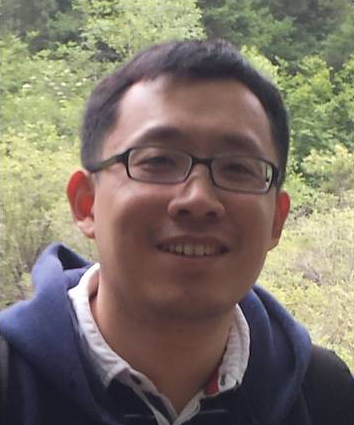}}]{Weiyao Lin}
received the B.E. and M.E. degrees from Shanghai Jiao Tong University, China, in 2003 and 2005, and the Ph.D degree from the University of Washington, Seattle, USA, in 2010.

He is currently an associate professor at Department of Electronic Engineering, Shanghai Jiao Tong University, China. His research interests include image/video processing, video surveillance, computer vision.
\end{IEEEbiography}

\vspace{-28pt}
\begin{IEEEbiography}[{\includegraphics[width=1in,height=1.2in,clip,keepaspectratio]{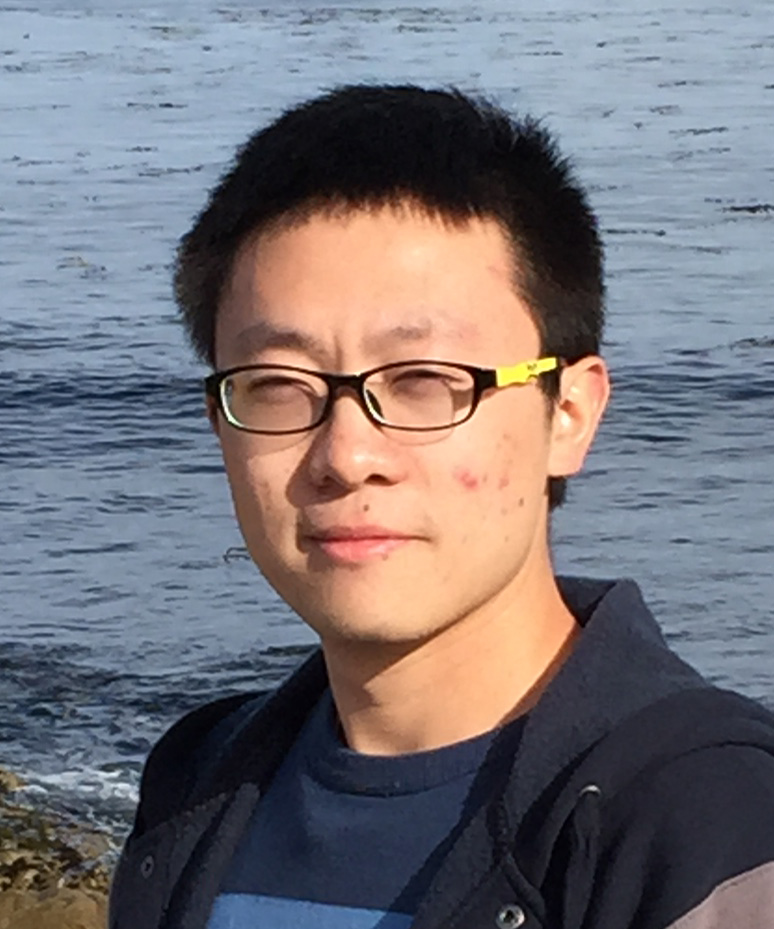}}]{Yang Zhou}
received the B.S. degree from Shanghai Jiao Tong University, Shanghai, China, in 2013 in Electrical Engineering. He is currently working toward the M.E. degree in Electrical Engineering from both Shanghai Jiao Tong University, China and Georgia Institute of Technology, USA. His research interests include image processing \& computer vision.
\end{IEEEbiography}

\vspace{-28pt}

\begin{IEEEbiography}[{\includegraphics[width=1in,height=1.2in,clip,keepaspectratio]{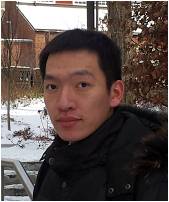}}]{Hongteng Xu}
received the B.S. degree from Tianjin University, Tianjin, China, in 2010. In fall 2010, he joined in the dual-master program of Georgia Institute of Technology and Shanghai Jiao Tong University, and graduated in Spring 2013. Currently, he is a Ph.D. student, School of Electrical and Computer Engineering, Georgia Tech. His research interests include computer vision, data mining and machine learning.
\end{IEEEbiography}

\vspace{-28pt}

\begin{IEEEbiography}[{\includegraphics[width=1in,height=1.2in,clip,keepaspectratio]{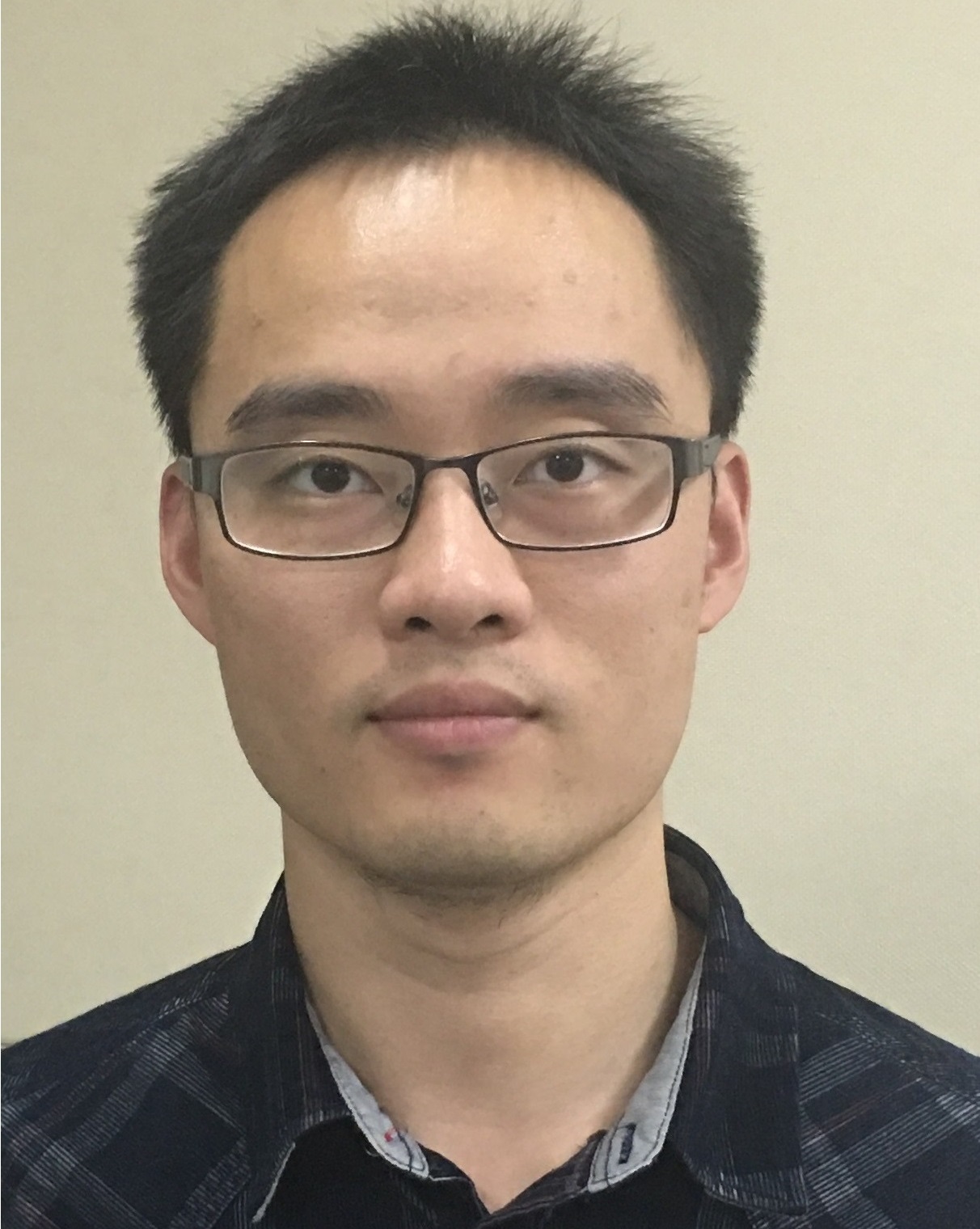}}]{Junchi Yan}
received the M.S. and Ph.D degrees from Shanghai Jiao Tong University, China. He is currently a Research Staff Member and Master Inventor with IBM.

He is the recipient of IBM Research Accomplishment and Outstanding Accomplishment Award in 2013, 2014. His research interests are computer vision and machine learning applications.
\end{IEEEbiography}

\vspace{-28pt}

\begin{IEEEbiography}[{\includegraphics[width=1in,height=1.2in,clip,keepaspectratio]{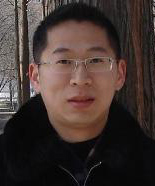}}]{Mingliang Xu}
received the B.E. and M.E. degree from Zhengzhou University, China, in 2005 and 2008, respectively, and the Ph.D degree from the State Key Lab of CAD\&CG at Zhejiang University, China, in 2012, all in computer science. Currently, he is an associate professor in School of Information Engineering of Zhengzhou University, China. His research interests include computer graphics \& computer vision.
\end{IEEEbiography}

\vspace{-28pt}

\begin{IEEEbiography}[{\includegraphics[width=1in,height=1.2in,clip]{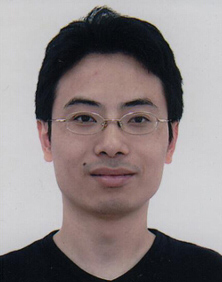}}]{Jianxin Wu}
received his B.S. \& M.S. degree in computer science from Nanjing University, and the PhD degree in computer science from Georgia Institute of Technology.

He is currently a professor in Nanjing University, China and was an assistant professor in Nanyang Technological University, Singapore. His research interests are computer vision and machine learning.
\end{IEEEbiography}

\vspace{-28pt}

\begin{IEEEbiography}[{\includegraphics[width=1in,height=1.2in,clip]{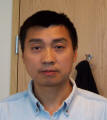}}]{Zicheng Liu}
is a principal researcher at Microsoft Research Redmond. He received his Ph.D. in computer science from Princeton University in 1996, his B.S. from Huazhong Normal University, China, in 1984, and M.S. from Institute of Applied Mathematics, Chinese Academy of Sciences in 1989. Before joining Microsoft Research, he worked at Silicon Graphics Inc.

His current research interests include human activity recognition, 3D face modelling, and multimedia signal processing. He is a fellow of IEEE.
\end{IEEEbiography}

\end{document}